\documentclass[a4paper,fleqn]{cas-dc}

\usepackage[numbers]{natbib}

\usepackage{soul}
\usepackage{url}
\usepackage{graphicx}
\usepackage{subcaption}
\usepackage{threeparttable} 
\usepackage{bbding}

\newcommand{\norm}[1]{\|#1\|}

\newcommand{\RN}[1]{%
  \textup{\uppercase\expandafter{\romannumeral#1}}%
}
\usepackage{amsmath} 
\usepackage{booktabs}
\usepackage{colortbl}

\graphicspath{{Fig/}}

\usepackage{amsthm}
\usepackage{lipsum} 

\newtheoremstyle{customdefinition}
  {\topsep}   
  {\topsep}   
  {\normalfont} 
  {}          
  {\bfseries} 
  {.}         
  {.5em}      
  {\thmname{#1}\thmnumber{\space#2}\thmnote{\space(#3)}} 

\theoremstyle{customdefinition}
\newtheorem{innercustomdefinition}{Definition}[section]
\newenvironment{customdefinition}[1]
  {\renewcommand\theinnercustomdefinition{#1}\innercustomdefinition}
  {\endinnercustomdefinition}

\definecolor{extralight}{RGB}{170, 237, 250}
\definecolor{light}{RGB}{29, 184, 210}
\definecolor{medium}{RGB}{22, 144, 169}
\definecolor{dark}{RGB}{7, 70, 94}

\usepackage{tikz}
\usetikzlibrary{fadings}
\usepackage{tikz}
\usetikzlibrary{arrows}
\usepackage{booktabs,caption}
\usetikzlibrary{plotmarks} 
\usetikzlibrary{patterns}
\usetikzlibrary{decorations.markings}
\usetikzlibrary{calc}
\usetikzlibrary{shapes}
\usetikzlibrary{shapes.geometric}
\usetikzlibrary{arrows.meta}
\usetikzlibrary{fit}

\usepackage{fancyhdr}

\begin{document}
\let\WriteBookmarks\relax
\def\floatpagepagefraction{1}
\def\textpagefraction{.001}

\shorttitle{Temporal dynamics of mutations to enhance HIV-1 therapy prediction}

\shortauthors{Di Teodoro et~al.}

\title [mode = title]{Incorporating temporal dynamics of mutations to enhance the prediction capability of antiretroviral therapy's outcome for HIV-1}                   



%
\author[1,2]{{Giulia} {Di Teodoro}}
\ead{diteodoro@diag.uniroma1.it}


\author[3,4]{Martin Pirkl}
\author[2,5]{Francesca Incardona}
\author[6]{Ilaria Vicenti}
\author[7,8]{Anders Sönnerborg}
\author[3,4]{Rolf Kaiser}
\author[1]{Laura Palagi}
\author[6]{Maurizio Zazzi}
\author[3,9]{Thomas Lengauer}

\affiliation[1]{organization={Sapienza University of Rome, Department of Computer Control and Management Engineering Antonio Ruberti}, postcode = {00185}, city={Rome}, country={Italy}}
\affiliation[2]{organization={EuResist Network}, postcode = {00152}, city={Rome}, country={Italy}}
\affiliation[3]{organization = {University of Cologne, Institute of Virology, Faculty of Medicine and University Hospital Cologne}, postcode={50935}, city={Cologne}, country={Germany}}
\affiliation[4]{organization={Partner Site Cologne-Bonn, German Center for Infection Research (DZIF)},postcode={50935}, city={Cologne}, country={Germany}}
\affiliation[6]{organization={University of Siena, Department of Medical Biotechnologies}, postcode={53100}, city={Siena}, country={Italy}}
\affiliation[5]{organization={I-PRO}, postcode = {00152},city={Rome},country={Italy}}
\affiliation[7]{organization ={Karolinska Institutet, Division of Infectious Diseases, Department of Medicine Huddinge}, postcode ={14152}, city={Stockholm}, country={Sweden}}
\affiliation[8]{organization={Karolinska University Hospital, Department of Infectious Diseases},postcode={14186}, city ={Stockholm}, country={Sweden}}
\affiliation[9]{organization={Max Planck Institute for Informatics, Saarland Informatics Campus, Computational Biology},postcode={66123}, city={Saarbrücken}, country={Germany}}

\begin{abstract}
\\ \textbf{Motivation:} In predicting HIV therapy outcomes, a critical clinical question is whether using historical
information can enhance predictive capabilities compared with current or latest available data analysis. This study analyses whether historical knowledge, which includes viral mutations detected in all genotypic tests before therapy, their temporal occurrence, and concomitant viral load measurements, can bring improvements. We introduce a method to weigh mutations, considering the previously enumerated factors and the reference mutation-drug Stanford resistance tables. We compare a model encompassing history (H) with one not using it (NH). \\
\textbf{Results:} The H-model demonstrates superior discriminative ability, with a higher ROC-AUC score (76.34\%) than the NH-model (74.98\%). Significant Wilcoxon test results confirm that incorporating historical information improves consistently predictive accuracy for treatment outcomes. The better performance of the H-model might be attributed to its consideration of latent HIV reservoirs, probably obtained when leveraging historical information. The findings emphasize the importance of temporal dynamics in mutations, offering insights into HIV infection complexities. 
However, our result also shows that prediction accuracy remains relatively high even when no historical information is available.\\
\textbf{Supplementary information:} Supplementary material is available.
\end{abstract}



\begin{keywords}
human immunodeficiency virus \sep antiretroviral drug \sep therapy prediction \sep rate of mutation disappearance \sep weighting factors for mutations \sep viral load \sep Stanford mutation-drug resistance score \sep machine learning 
\end{keywords}

\maketitle

\pagestyle{fancy}
\lhead{}
\renewcommand{\headrulewidth}{0pt}
\rhead{\textcolor[gray]{0.3}{This version has been published in Bioinformatics: \href{https://doi.org/10.1093/bioinformatics/btae327}{https://doi.org/10.1093/bioinformatics/btae327}}}
\thispagestyle{fancy}  

\textcopyright{ 2024. Licensed under the Creative Commons  \href{https://creativecommons.org/licenses/by-nc-nd/4.0/}{CC-BY-NC-ND 4.0}.} 

\section{Introduction}
Human immunodeficiency virus (HIV), if untreated, is a deadly pathogen. While there are treatment options, there is still no cure or vaccine. Since its discovery in 1981, 84.2 [64.0-113.0] million people have been infected with HIV, claiming about 40.1 [33.6-48.6] million lives. By the end of 2022, about 39.0 million people were living with HIV \citep{WHO}. Early diagnosis and proper treatment offer life expectancy comparable to HIV-negative individuals.
HIV-1 infection requires antiretroviral treatment; without it, patients eventually develop acquired immunodeficiency syndrome (AIDS). Standard care involves the administration of cocktails of antiretroviral drugs, rather than one individual medicine, to minimize the risk of emergent drug resistance. Antiretroviral therapies against HIV-1 can lead to the selection of drug-resistant HIV-1 variants that can spread between hosts \citep{vanDerKlundert_2022}, causing treatment failure \citep{Langford2007-ot,Wensing2022-iw}. Resistance testing can help to suppress viral replication and to prevent the transmission of resistant variants. The analysis of the susceptibility of the HIV-1 variants to available antiretroviral drugs can be facilitated via genotypic testing or phenotypic testing \citep{Tang2012-zy}. Phenotypic testing directly measures the drug concentration required to inhibit virus replication \textit{in vitro}, whereas genotypic testing comprises sequencing the viral genome and inferring drug susceptibility based on prior knowledge of resistance mutations. Genotypic resistance tests, being the more practical variant, are routinely used in clinical practice. If drug-resistant viral variants emerge, therapy must be replaced with a different drug combination to suppress replication.
Clinicians struggle with identifying new drug combinations to suppress HIV replication, taking into account viral drug resistance, previous successful and failing therapies, and retention of future treatment options.
However, the latent viral population in the patient living with HIV (PLWH) accumulates a large number of resistance-associated mutations, which may not be observable in blood serum but rapidly reappear when it is advantageous, i.e., under the appropriate drug pressure \citep{Rhee22,Wensing2022-iw}. The many possible drug combinations complicate therapy selection, especially in advanced stages. For this reason, over the years genotypic interpretation systems for drug resistance have been developed for predicting the success or failure of antiretroviral drug regimens.\\
Genotypic drug resistance interpretation systems (GIS) are rules-based or data-driven interpretation systems.
The former approach uses tables of drug-resistance mutations assembled by expert groups \citep{Wensing2022-cr} to calculate drug-resistance scores that assess drug resistance of viral genotypes. Several rules sets have been developed over the years, such as the ones from ANRS, HIVdb \citep{Tang2012-po}, HIV-GRADE, and the Rega Institute, all available on the HIV-GRADE website \citep{Obermeier2012-pu}. These rules-based systems are continually updated according to changes in observed HIV drug resistance, treatment guidelines, and expert opinion \citep{Paredes2017-gm}. All of these systems interpret data obtained from Sanger sequencing, a long-standing technology widely used thanks to its low rate of error and its cost-effectiveness. However, this technique has limits in detecting minor resistant populations \citep{Davidson2012-pl,Tsiatis2010-bj} that could have clinical relevance \citep{Cozzi-Lepri2015-wo,Vrancken2016-oc}. High-throughput sequencing techniques, collectively referred to as next-generation sequencing (NGS) \citep{Fox2014-rg}, are now being increasingly used which can detect minority variants representing as low as 1\% of the viral population, as opposed to roughly 20\% with Sanger sequencing. The mutational patterns detected by NGS are currently subjected to the same genotype interpretation systems as those detected by Sanger sequencing. However, the clinical role of drug resistance mutations between 1\% and 20\% prevalence is still a matter of debate and is likely different for different drugs and mutations. 
On the other hand, data-driven genotypic interpretation systems rely on statistical or machine-learning (ML) methods to infer drug resistance directly from data. Due to the large amount of clinical and genotypic data available, data-driven GIS have become a prominent approach to help clinicians choose effective HIV therapies, especially for heavily treatment-experienced patients with complex drug resistance patterns that have evolved over years. Early data-driven GISs, like the Virco proprietary VirtualPhenotype$^{\textit{TM}}$ \citep{Vermeiren2007-xp} and the geno2pheno system \citep{Beerenwinkel2003-jf}, predict the \textit{in-vitro} phenotype. The Virco system (VircoTYPE) was initially based on a linear regression model that estimates the phenotypic measurement as the weighted sum of the effects of individual mutations and then adapte so as to transform the predicted phenotype into clinically relevant estimate of efficacy \citep{Winters2008-nj}. Geno2pheno initially assessed virus resistance to individual compounds (Geno2pheno[resistance]) and later predicted virological responses to antiretroviral (ART) regimen comprising up to four drugs. Geno2pheno-THEO estimated the probability of treatment success using genotypic data and user inputs \citep{Altmann2009-vu}. Geno2pheno[resistance] \citep{Lengauer2006-wm,Pironti2017-wi} uses data on viral sequence and on drugs for classification via the ML technique called support-vector machine (SVM). Geno2pheno[drug exposure] \citep{Pironti2017} uses the same input first to produce so-called drug exposure scores (DESs) correlated with prior drug exposure and secondly to supply the probabilities of exposure derived from the produced DESs to a statistical model for therapy prediction. 
Another software developed to infer the virological response to antiretroviral drug regimen is SHIVA which employs random forests \citep{Riemenschneider2016SHIVAA}. In one study, artificial neural networks were used for the same purpose \citep{Larder2007-nx}. \\
Some research suggests that mutations that have been observed in the past and then disappeared from blood serum can impact a patient's current status and the effectiveness of subsequent therapy. The time since a mutation first appeared or disappeared from blood serum may influence the degree to which that mutation informs on a patient's status and resistance to antiretroviral drugs \citep{Ciccullo2021-er,Gagliardini2018-ud}. In addition, the impact of a mutation is assumed to be more important the higher the viral load (VL) when the mutation was observed. 
In light of these considerations, in our model, we consider not only the mutations detected by the patient's most recent genotypic resistance test (GRT), as in models currently in routine use. Rather, we consider the entire history of mutations recorded for the patient in the database before onset of the therapy of interest, for which we want to correctly predict the success or failure for that particular patient. Mutations are assumed to contribute additively to the patient's resistance status. The contribution of each mutation is multiplied by a weight that incorporates \textit{(i)} a degression factor for time: the further back in time the mutation was observed, the less likely it is to be still influential in determining the patient's drug resistance, \textit{(ii)} the area under the VL curve measured in a time window around the date of the mutation's occurrence: the larger that area the more informative the mutation is considered to be, \textit{(iii)} a penalty score based on the HIVdb system, in which each drug resistance mutation (DRM) is assigned a drug penalty score based on medical expertise. This score is widely recognized in the literature. Since the latent virus in organ tissue is not accessible to routine diagnostics, with this methodology, we aim to infer hidden mutations from history data on mutations  detected in blood serum. Moreover, our system is based on a single support vector machines (SVM) model that predicts the therapy outcome from the patient's mutation history.

\section{Methods}
\subsection{Problem setting}\label{sec:problem}
Let $\mathbf{M} = \{1,..., M\} $ ($M=5941$) be the set of mutations considered, $m\in\mathbf{M}$, denote a specific mutation, and $date_m$ the most recent point in time when that mutation was recorded in a GRT. Let $\mathbf{D}= \{1,...,D\}$ be the set of drugs considered and $d \in\mathbf{D}$ denote an individual drug. Let $r\in \{0,1\}^M$ be a vector indicating all DRMs observed in at least one viral genotype in the set of viral genotypes considered before the onset of the therapy of interest. Eventually, the mutations this vector indicates ($r_i=1$) will be weighted with the weighting factors explained in the section \ref{sec:weights}. Let $z\in \{0,1\}^D$ be a vector indicating the combination of drugs used in a particular therapy of interest, and let $start_z$ denote the time point of the onset of therapy $z$. 
Label $y\in\{0,1\}$ indicates success (0) or failure (1) of the therapy. Success and failure of therapy are defined as denoted in the definition of the Standard Datum, which is given in the section \ref{section: The dataset}.
The model is therapy-oriented, such each datapoint in the training set is composed of the $i^{th}$ \textit{patient-therapy} pair $x_i =(r_i,z_i)$  and the corresponding value of the output $y_i$ indicating efficacy of therapy; hence it is a triple $(r_i,z_i,y_i)$.
Let $\mathbf{T} = $\{$(r_{1},z_{1},y_{1}),...,(r_{N},z_{N},y_{N})\}$ be the training set, where $N$ is the cardinality of the set of patient-therapy pairs. Our goal is to train several models $f(x)$ based on SVM that accurately predict the outcome of a target therapy $z$.The models differ in that some of them take information on the medical history of the patient into account and others do not. A comparison of the performances of models will afford insights on how relevant therapy history is for therapy prediction. 

\subsection{The weighting factors of the mutations}\label{sec:weights}
\begin{figure*}[!t]
  \centering
  \begin{subfigure}[b]{0.32\textwidth}
    \includegraphics[width=\textwidth, height=3.5cm]{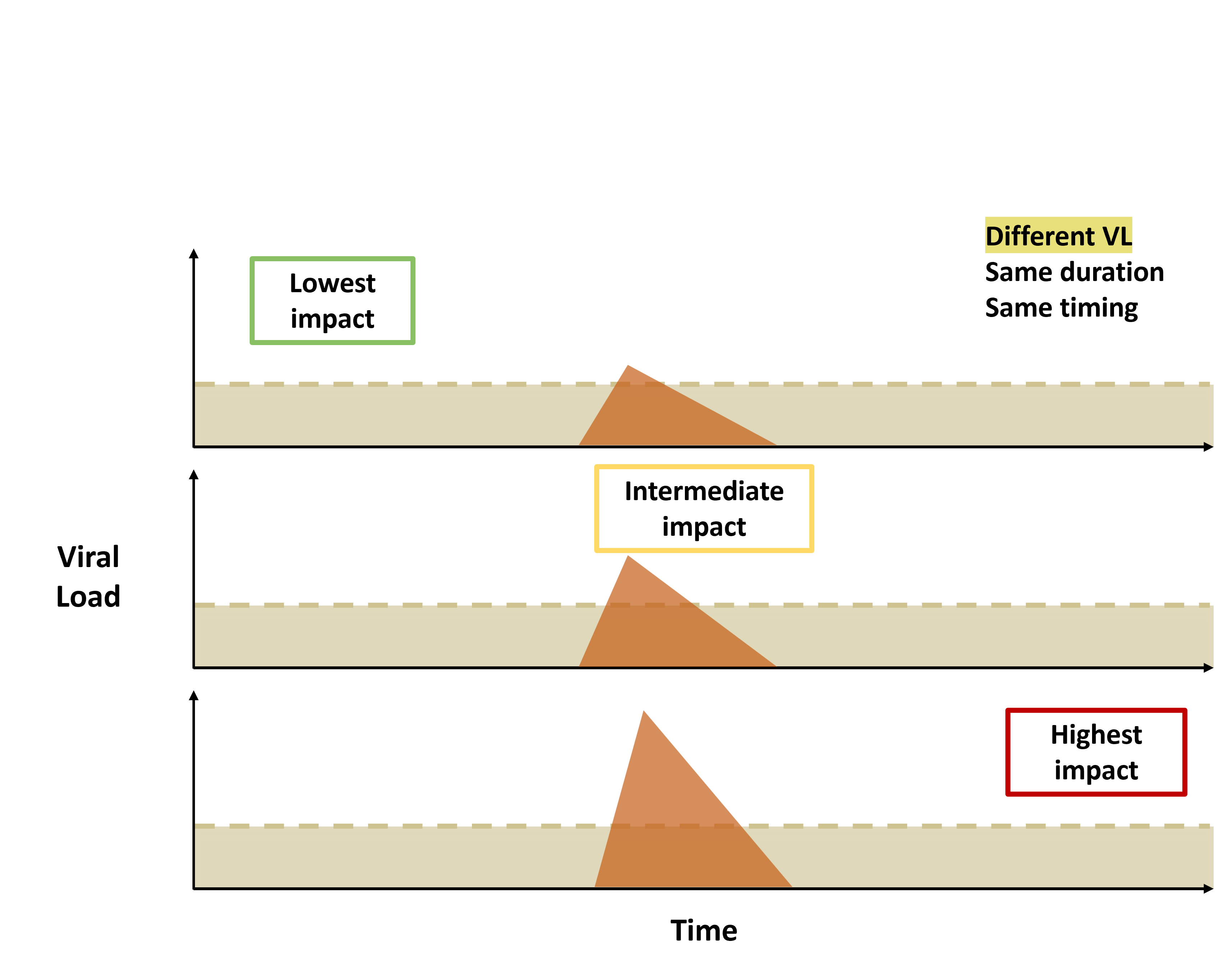}
    \caption{}
    \label{fig:sub1_imp}
  \end{subfigure}
  \hfill
\begin{subfigure}[b]{0.32\textwidth}
    \includegraphics[width=\textwidth, height=3.5cm]{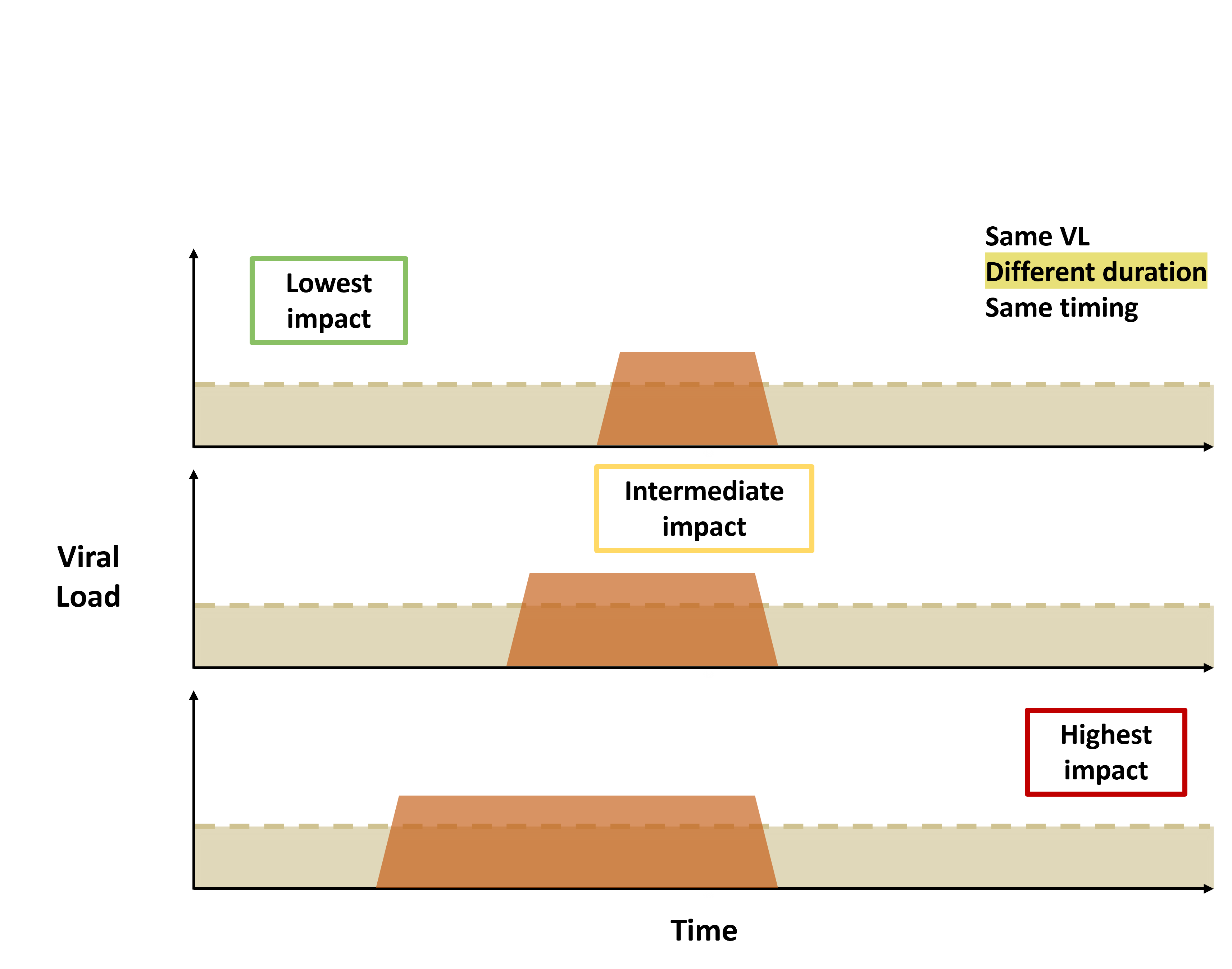}
    \caption{}
    \label{fig:sub2_imp}
  \end{subfigure}
 \begin{subfigure}[b]{0.32\textwidth}
\includegraphics[width=\textwidth, height=3.5cm]{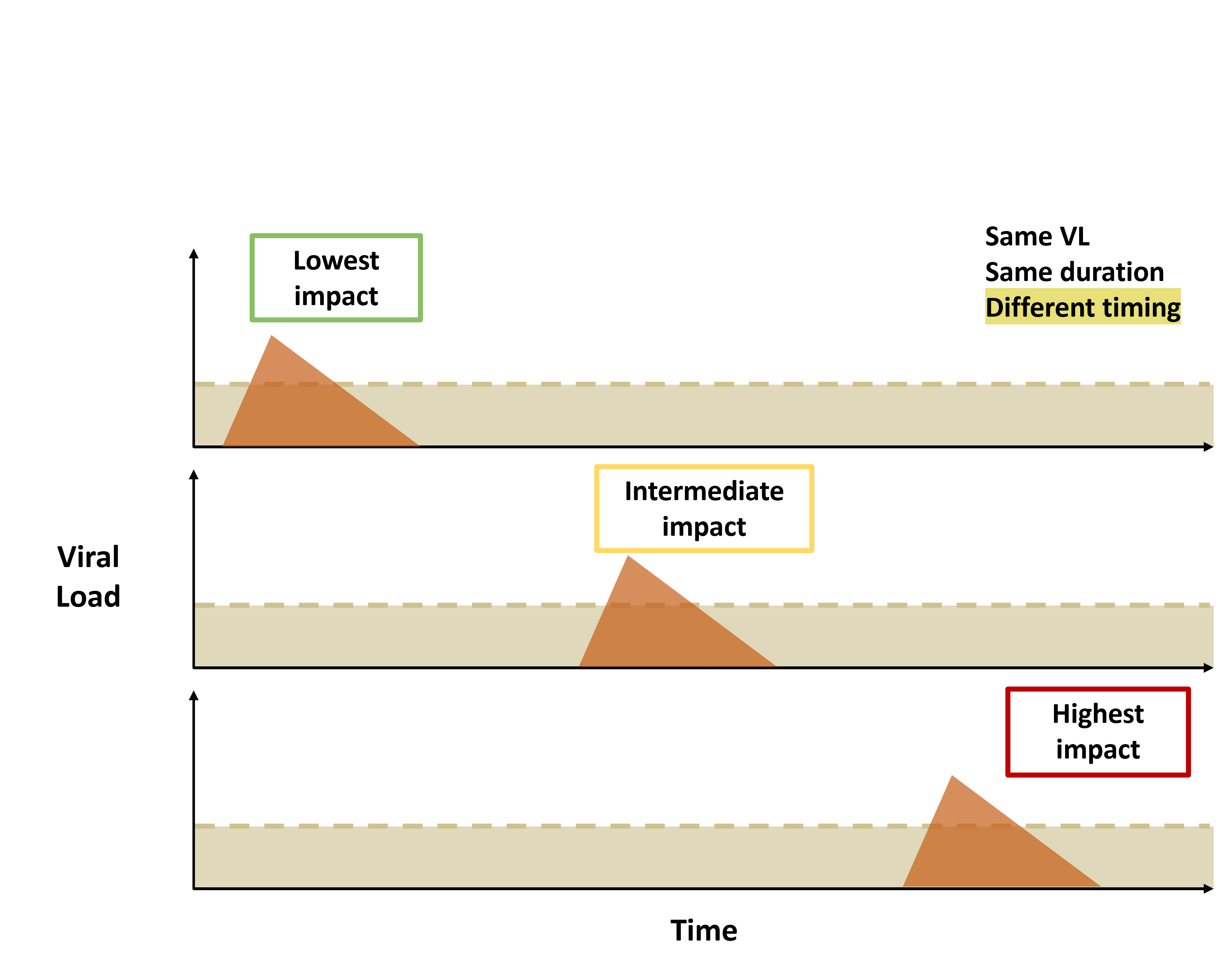}
\caption{}
    \label{fig:sub3_imp}
  \end{subfigure}
\caption{Figure (a) represents how different viral loads, measured when a mutation was detected, can impact present drug resistance. Figure (b) represents how different durations of a mutation can impact present drug resistance. Figure (c) represents how different mutation timing can impact present drug resistance.}
  \label{fig:impact}
\end{figure*}

Previous experience suggests that the timing and duration of mutations could affect therapy efficacy. In addition, the magnitude of VL when a mutation is detected could influence viral replication. How these factors could impact the present drug resistance is shown schematically in Figure \ref{fig:impact}. This is precisely why these factors are incorporated in the calculation of the weight factor for a mutation. In addition, the Stanford Score table is also taken into account. This is a table of scores indicating the impact of individual mutations on drug resistance of the virus. The scores for the mutations observed in a viral genotype are accumulated to yield an overall score quantifying the drug resistance of the virus to an ART regimen. The Stanford Score table is widely used as an independent predictor of virologic response to drug regimens \citep{Tang2012-po}. \\

\begin{figure}[!t]
  \centering
  \begin{subfigure}[b]{0.42\columnwidth}
    \includegraphics[width=\columnwidth, height=2.7cm]{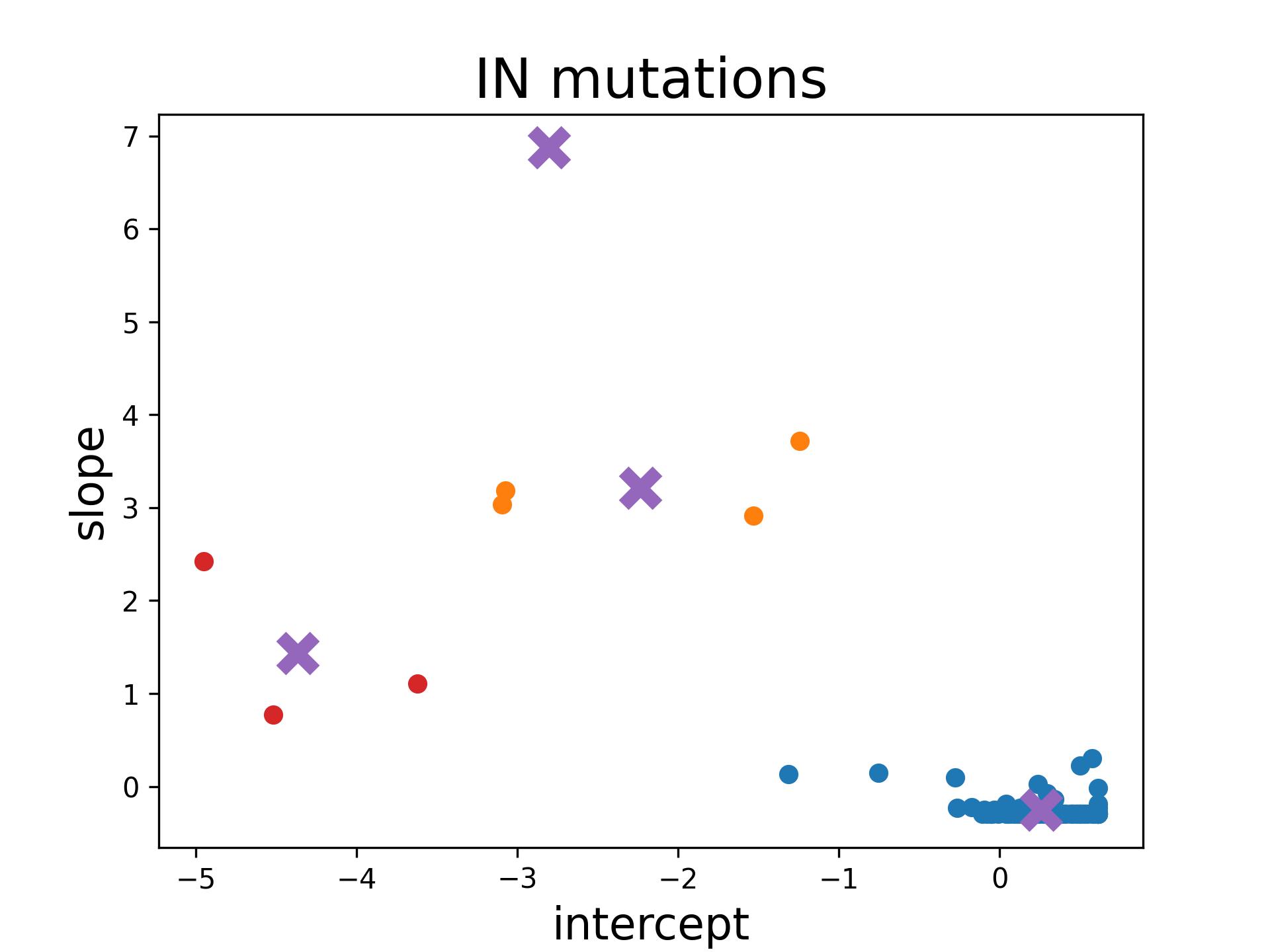}
    \caption{}
    \label{fig:sub1_si}
  \end{subfigure}
  \hfill
\begin{subfigure}[b]{0.42\columnwidth}
    \includegraphics[width=\columnwidth, height=2.7cm]{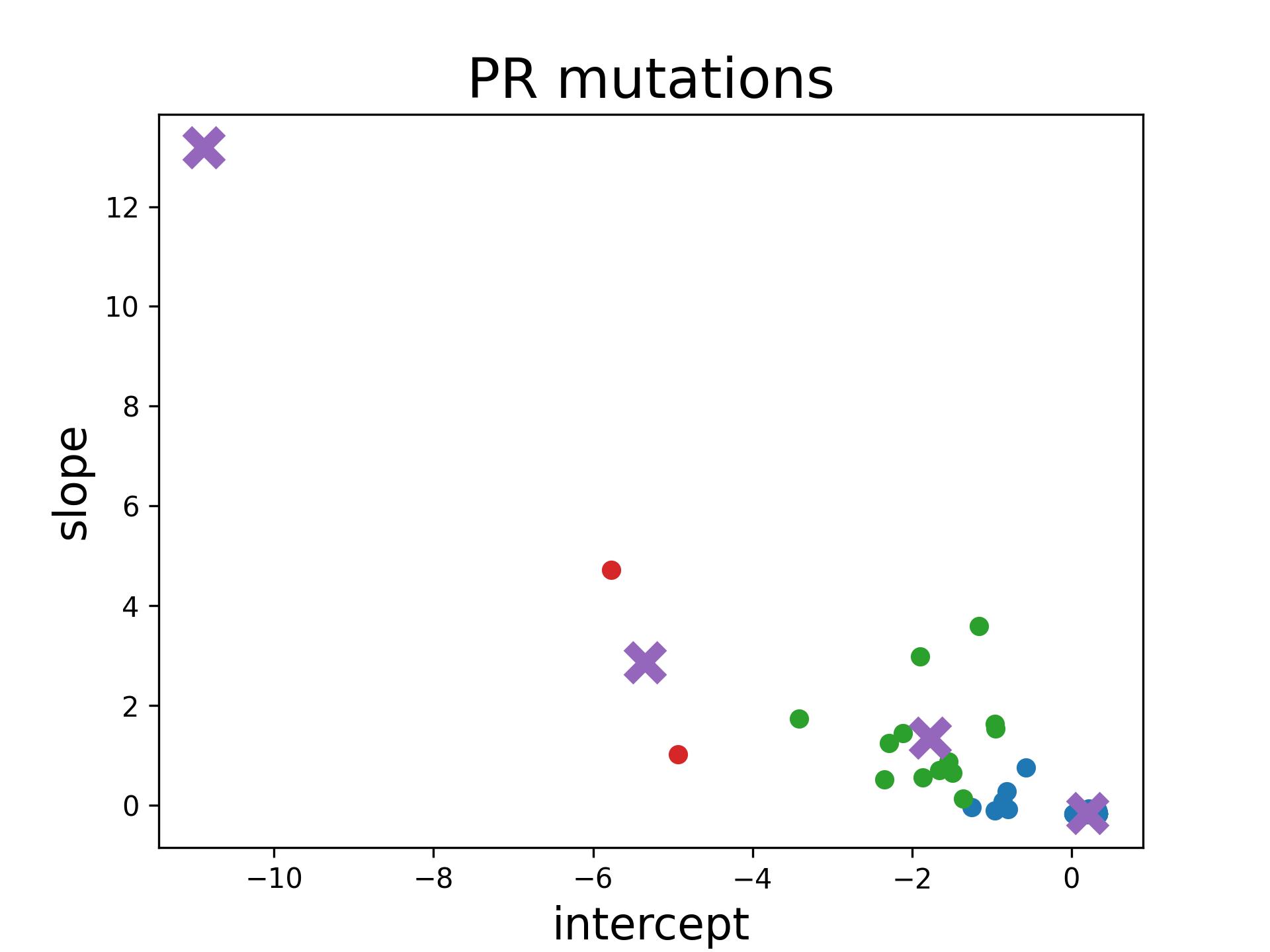}
    \caption{}
    \label{fig:sub2_si}
  \end{subfigure}
 \begin{subfigure}[b]{0.42\columnwidth}
\includegraphics[width=\columnwidth, height=2.7cm]{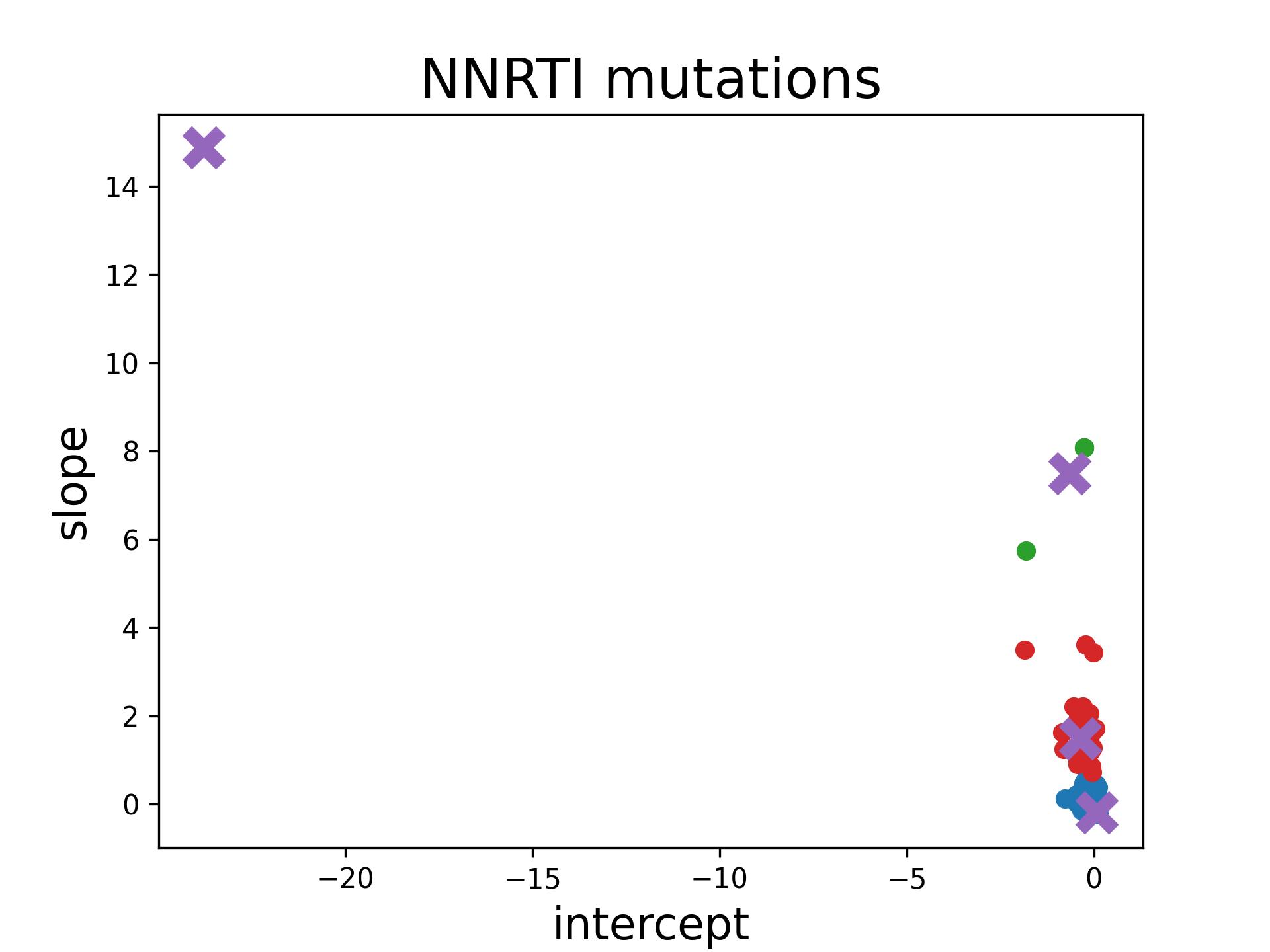}
\caption{}
    \label{fig:sub3_si}
  \end{subfigure}
\caption{Figure (a) slope-intercept pairs clustered for IN-mutations, Figure (b) slope-intercept pairs clustered for PR-mutations and Figure (c) slope-intercept pairs clustered for NNRTI-mutations. }
  \label{fig:ranking_coef}
\end{figure}

\textbf{Time}. The closer the most recent time point that a mutation has been observed lies to the initiation of a drug regimen, the greater its impact on the efficacy of this regimen is assumed to be. The reason is that the latent reservoir of provirus in tissues is assumed to deplete over time. We modeled the rate of disappearance of a mutation via a sigmoid curve: for a certain period of time, the mutation has a high propensity of impacting drug resistance, then this propensity begins to decline, tending to zero. 
To learn the onset of the decline of the curve $\alpha_{m}$ and slope $\beta_{m}$ of each sigmoid curve $y(t)=\frac{1}{1+e^{\alpha+\beta t}}$ pertaining to an individual mutation, the data on when mutations were and were not observed, respectively, were collected as explained below. \\
The mutations were grouped by drug class (Protease Inhibitor (PI), Nucleoside Reverse Transcriptase Inhibitor (NRTI), Non-Nucleoside Reverse Transcriptase Inhibitor (NNRTI), Integrase inhibitor (INI)). To assess the rate of disappearance of a mutation, one has to consider periods when the patient is not on ART or is not taking any drug of a given drug class that might select for that mutation. \\
Consider, for example, mutations in the PI drug class. 
Starting with the entire database, we consider all patients who had a therapy with at least one drug of the PI class (PI therapy) and immediately after a period when they were on another therapy that contained no drug of the PI class or a period where the patient was not under any therapy so that there was no drug pressure due to any drug from the PI class in the patient. 
If we consider all PI mutations present in the viral sequences during the therapy, including one or more PIs, we can see how many of those mutations disappeared at some point in one of the sequences that were sampled subsequently. If a mutation was present during the PI therapy, the following cases might occur:
\begin{itemize}
    \item In the first sequence sampled after the PI therapy was stopped, the mutation has already disappeared. The mutation's persistence period is assumed to be the time between the date the PI therapy was stopped and the date this sequence was sampled.
\item The mutation is still present in the first sequence after discontinuing the PI therapy but disappears in the $n^{th}$ sequence observed. Here, the persistence period is the time between the date of discontinuation of PI therapy and the date of sampling of the  $n^{th}$  sequence.
\item The mutation is present in all subsequent sequences until the end of the period in which the patient is off any PI drug, thus we consider that the mutation has not disappeared.
\end{itemize}
Therefore, for the $m^{th}$ mutation, we construct two vectors:
\begin{itemize}
    \item The vector $x_m$ of days elapsed from the date of discontinuation of the PI therapy to the sampling date of gene sequences at a later time point when the patient experiences no drug pressure due to drugs in the PI class (e.g., $x_{m} = [20, 50, 250, 347, 500, 1000]$ would represent a sequence of six consecutive GRTs that occur on the indicated number of days after stopping a PI therapy, i.e., the first test happens on day 20, the second on day 50, etc).
    \item The vector of binary values indicating whether, at each time point indicated by the vector $x_m$, the mutation is present (1) or has not been observed (0). (e.g., $y_{m} = [ 1, 1 , 1 , 1 , 0 , 0 ]$, corresponding to the vector of the previous example, indicates that the $m^{th}$-mutation is present in the first 4 GRTs performed after therapy stop, but in the fifth test, after 500 days, is no longer there. In the data, there are also situations in which, e.g., $y_{m} = [ 1, 1 , 1 , 0 , 1 , 0 ]$ where the mutation reappears.
 \end{itemize}
Let $x = (x_{1}, ..., x_{n})$  be a random sample of $n$ observations from the distribution with \textit{probability distribution function (pdf)} $ f(x; \theta) = \frac{1}{1+e^{(\alpha+\beta x)}}$ depending on the model parameter $\theta$$=(\alpha;\beta)$. We defined the target probabilities $t_{m} = \frac{y_{m}+1}{2}$ and minimized the negative log-likelihood function \citep{Platt1999} $$ \min_\theta -\frac{1}{n}\sum(t\log(f(x; \theta)) +(1-t)\log(1-f(x; \theta)))$$ \\
Equipping our statistical model for therapy efficacy with individual degression curves for each mutation is not promising, given the small amount of data we have. Thus, we decided to cluster the slope-intercept pairs of the curves for mutations pertaining to a drug class with a \textit{k-means} algorithm after standardizing them using the \textit{z-score}.\\ 
This procedure was repeated for mutations pertaining to all drug classes, except for the NRTI class, because drugs from this class have been administered practically always without interruption. The clusters are depicted in Figures \ref{fig:sub1_si}, \ref{fig:sub2_si} and \ref{fig:sub3_si}, where the points represent the slope-intercept pairs, the colors refer to the different clusters and the crosses represent the clusters' centroids. For mutations for which no data were available to learn the slope and intercept of the sigmoid functions, the slope and the intercept of their sigmoid curves are treated as hyper-parameters of the model to be optimized by cross validation.\\

\textbf{Viral load}. The results in previous studies \citep{Liu2022-zq} suggest that the viral load observed in the presence of a mutation could influence the impact that the mutation has on drug resistance. To take this into account, since a value of the VL is not always present for the same day that the genetic sequence was sampled for GRT, it was decided to consider a two-month time window around the sample date and to use the area under the viral load curve obtained by performing linear interpolation of the viral load values measured within that time window. We will refer to this area as $Area_{\textup{VL}}$. Note that the viral load values were logarithmized and decreased by log(50) such that VLs values below 50\textit{cp/ml} contribute a negative area. The rationale is that 50 is the threshold of undetectability currently used in clinical routine. For each mutation detected in the genotypic sequence of different patients, the values of $Area_{\textup{VL}}$ were normalized to the interval $[-1.1]$. \\ 

\textbf{The Stanford Score}. The Stanford score is associated with a mutation-drug pair, $s_{m,d}$.  This score is the higher the more a given mutation decreases the susceptibility of the virus to the drug. Not all mutations are considered in the Stanford tables; therefore, for those for which this information is not available, the score is set to 0. 
If a mutation is associated with multiple drugs (across drug classes) included in the drug regimen $z_{i}$ and therefore multiple Stanford scores apply, the minimum Stanford score is used, because the most effective drug can be assumed to dominate the effect on the virus. \\
To fend against biases introduced by learning on largely differing values all scores were standardized by dividing them by the norm of the vector of all possible Stanford score $s = [-15, -10, -5, 0, 5, 10, 15, 20, 25, 30, 35, 40, 45, 50, 55, 60]$. Hence, the Stanford score associated with a mutation is defined as follows: $$S_{m} = \frac{ \min \{ s_{m,1},...,s_{m,d},..., s_{m,D}\}}{\norm{s}} , \mbox{ with } d\in h$$  
The weight to be given to mutation $m$ is defined as follows. 
\begin{equation} \label{eq_ wight_mutation}
\centering
    w_{m} = \frac{Area_{\textup{VL}} }{1+e^{\alpha+\beta t} - \tanh(S_{m})}
\end{equation}
where \textit{t} is the time passed (in days) between the last time the mutation $m$ was detected and the start of the therapy of interest $z$. I.e., an increase in viral load increases the maximum weight of the mutation, and the Stanford score decreases the effect of fast and steep descent of the weight. For each mutation $m$, $w_m \in [-1,1] $.\\
To evaluate the efficacy of this approach, we trained multiple linear SVM models, contrasting the effectiveness of incorporating the patient's complete mutational profile with baseline genotype analysis alone. In this context, we could use simple binary values or mutation-specific weights – calculated as in \ref{eq_ wight_mutation} – for representing mutations in each patient-therapy sample to be given as input to the model. The choice to use linear models was dictated by the desire to maintain the interpretability of the results.

\section{Experiments and results}
\subsection{The EuResist integrated database}
The Euresist Integrated Database EIDB is one of the world's largest databases regarding drug resistance in HIV-infected patients, both treatment-naïve and treated, undergoing clinical follow-up since 1998 \citep{Rossetti}. The data comprises nine national cohorts from Italy, Germany, Sweden, Portugal, Spain, Luxembourg, Belgium, Turkey, and Russia. Recently, data from patients referred to facilities in Ukraine and Georgia have been added. The EuResist Database was established in 2006, to collect, in an anonymized form, data on demographic and clinical characteristics of PLWH, such as antiretroviral therapies, reasons for changing therapy, treatment responses, CD4+ cell counts, AIDS-defining events, and viral co-infections. \\
We trained several models for predicting the success or failure of antiretroviral treatments for patients who could be either treatment-naïve or already treated. The predictors in the model include \textit{(i)} the cumulative sequence of the predominant viral strains in patients' blood collected by all genotypic resistance tests (GRTs) performed before the start of the therapy of interest, \textit{(ii)} information on mutations as downloaded from the Stanford HIVDB, \textit{(iii)} viral load (VL) measurements as viral RNA copies per \textit{ml} of blood plasma (\textit{cp/ml}), and \textit{(iv)} the individual drugs used in the therapies. The response is the therapy outcome. It should be emphasized that the clinical data available on a patient is not necessarily complete. 

\subsection{The dataset} \label{section: The dataset}
The database contains data from $105,101$ PLWH but, for many of these patients, no consistent information is available to be used for our analysis. For example, nearly $8\%$ of patients $(8,346)$ do not have data on VLs, GRTs, or therapies with a valid date. Thus, in order to assemble the dataset for model training and testing, this database is preprocessed as described subsequently. Specifically, records on a subset of the patients in the dataset are selected for the analysis.\\
The models  are \textit{therapy-oriented}, which means that we organize the data in terms of \textit{patient-therapy pairs}. 
In the context of analyzing treatment success or failure, the notion of the tuples \textit{patient-treatment episode} (PTE) and \textit{patient-treatment change episode} (PTCE) has been introduced \citep{Zazzi2011-pg}. 
\begin{customdefinition}{Patient-treatment episode}
A patient-treatment episode (PTE) consists of a genotype (amino-acid sequence of reverse transcriptase (RT), protease (PR)and/or integrase (IN)) at baseline, the set of pharmacologic compounds used in antiretroviral treatment, (cART), an optional VL at baseline, obtained no earlier than 90 days before treatment initiation, and follow-up VLs, referred to a patient. Patient-treatment episodes include both patients at first-line therapies and patient-treatment change episodes.\end{customdefinition}
\begin{customdefinition}{Patient-treatment change episode}
     A patient treatment change episode (PTCE) is a type of PTE. It refers to a period during which data are collected to assess how the patient responds to the change in ART that has become necessary for some reason such as virologic failure, toxicity, drug interactions, or simplification of therapy. The start of the new treatment regimen serves as the "baseline," and the period considered is divided into two blocks, before and after start of treatment. During this episode, it is necessary to closely monitor the patient's HIV viral load and genotype at baseline and in the follow-up period and any possible side effects.
\end{customdefinition} 
Response to drug treatment is indicated with an outcome label  $y \in \{0,1\}$ indicating success or failure, respectively, according to a new EuResist Standard Datum definition that differs from the one used in the past. 
\begin{customdefinition}{Standard Datum} \label{def:standard Datum}
    Treatment success can be determined with a follow-up VL and optionally a VL at baseline as described. Follow-up VLs between 20 and 28 weeks after the start of therapy and the VL whose measurement date is closer to twenty-four weeks after the start of therapy are considered. Below, PTEs are referred to as successes, if the respective follow-up VL is less than 50 copies of HIV-1 RNA per milliliter of blood plasma. Otherwise, treatment is considered a failure. Cases in which treatment was changed before 20 weeks are considered as follows: \begin{itemize}
        \item Therapy lasting at most 4 weeks: excluded because most likely discontinued due to toxicity;  
        \item Therapy lasting 4 to 8 weeks: success if a viral load below 50 cps/ml or at least 1 log decrease in the last viral load before therapy stop was observed compared with the baseline test, otherwise failure;
        \item Therapy lasting 8 to 20 weeks: success if a viral load below 50 cps./ml or at least 2 log decrease in the last viral load before therapy stop was observed compared with the baseline test, otherwise, failure. \end{itemize}
\end{customdefinition}

A graphical representation of the Standard Datum is provided in Figure 1 of the Supplementary material.
The measurements of viral load have become more sensitive in recent years. Lower thresholds than previously can now be used to determine therapy success \citep{Zazzi2011-pg,Zazzi2012-ik}. Thus changing the Standard Datum accordingly in comparison with previous definitions is consistent with the current clinical practices.

In order to be included in the dataset, a patient-therapy pair must meet the following criteria: 
\begin{enumerate}
    \item The patient-therapy pair must comply with the definition of PTE, in particular, the full list of the compounds used in the therapy, at least one viral sequence observed before the start date of therapy and the follow-up VL must be present. 
    \item The patient that is administered that therapy must have at least one VL recorded before and after each GRT documented by a sequence at any temporal distance so that VL interpolation curves can be constructed. 
    \item The patient's therapy must be able to be classified successful or unsuccessful based on the definition of the standard datum.
\end{enumerate}
For each data point (patient-therapy pair), viral genotype information is provided in terms of a binary vector indicating the presence (1) or absence (0) of mutations, eventually multiplied by weights to account for the patient's history, as described in the section \ref{sec:weights}. The therapy to be administered next is also encoded by a binary vector indicating the presence or absence of the drugs appearing in the dataset. The full list of drugs considered is provided in the supplementary material.  \\
We consider both polymorphic and non-polymorphic mutations in our model. Polymorphisms are mutations occurring in at least 1\% of viruses not exposed to selective drug pressure, i.e., reflecting natural diversity independent from therapy. A nonpolymorphic mutation does not occur in the absence of therapy \citep{Shafer2007-qr}.
Our decision to consider all mutations is due to the fact that there may be mutations or combinations thereof that can result in reduced susceptibility of the virus to a cART or act on the fitness of the virus, not yet recognized as such.

In the end, our dataset consists of 22,000 therapy-patient pairs, among them 12,386 successes and 9,614 failures. We will refer to this dataset with the adjective \textit{Full}. Within the \textit{Full} dataset, we distinguish between two categories of samples: on the one hand, there are what we call type-1 therapy-patient pairs for which a detailed mutation history is available, collected through various GRTs performed prior to the start of the therapy of interest; on the other hand, there are type-2 pairs for which such a history is not available, either because they are first-line therapy or because the prior information is missing in the EIDB.\\
If there are no history data on a patient we also cannot leverage such data. Thus the type-2 pairs dilute our dataset in terms of the purpose of analyzing the worth of history information for therapy prediction. To address this issue we also consider what we call the \textit{Partial} dataset which comprises type-1 pairs exclusively. The \textit{Partial} database contains 10,581 \textit{patient-therapy} pairs (5,415 successes and 5,166 failures).
 \\

From each of these two datasets – \textit{Full} and \textit{Partial} – two different variants were constructed, differing by the choice of which mutations to consider for each data point:
\begin{itemize}
\item With-History Datasets: These datasets include all mutations identified in any GRT before the therapy under consideration, allowing the complete mutation history of a patient to be exploited for analysis. These datasets are referred to with the adjective \textit{History}.
  \item  Without-History Datasets: These sets are limited to the mutations detected in the last GRT performed prior to therapy. These datasets are identified by the adjective \textit{No-History}.
\end{itemize}

In summary, we have developed four variants of datasets- \textit{Full\_History}, \textit{Full\_No-history}, \textit{Partial\_History} and \textit{Partial\_No-history}-each designed for the purpose of exploring the influence of historical mutation information in treatment outcome prediction.

\subsection{The trained models}

\begin{figure}[t]
\centering
\includegraphics[width=9cm, height=2cm]{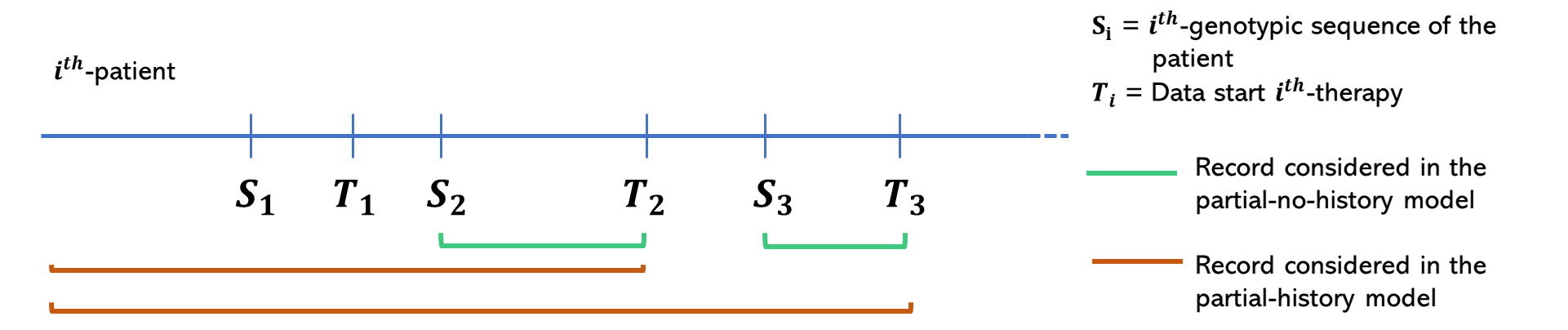}
\caption{Therapies and relative patient history considered in the history model and in the no-history model}
\label{fig:therapy-patient hist and no hist model}
\end{figure}

For each dataset, two different linear-SVM models were trained, as follows :
\begin{itemize}
\item \textit{Non-weighted} models: mutations are treated in a binary fashion. i.e., present (1) or absent (0).
\item \textit{Weighted} models: mutations are not represented in binary but weighted with the degression weight introduced above, offering varying degrees of significance or prevalence of mutations.
\end{itemize}
The models trained will be referred to as \textit{(Full/Partial)\_(History/ No-history)\_(Weighted/Non-weighted)} models.
The qualifiers \textit{History}, \textit{No-history}, \textit{Weighted} and \textit{Non-weighted} refer to the type of dataset used for training and testing and to the choice to weight or not to weight the mutations. Table 2 of the supplementary material presents a schematic view of the types of models we trained.
Figure \ref{fig:therapy-patient hist and no hist model} illustrates an example of the therapies considered in the \textit{Partial\_History} and \textit{Partial\_No-history} models, respectively. Therapy $T_{1}$ is not considered because prior to that therapy we have only one genotype at baseline. 


\subsection{Experimental setting}
The \textit{Full} datasets are slightly unbalanced: 56.3\% of the records pertain to successes. For the \textit{Partial} datasets, the respective fraction is 51.17\%. \\
The \textit{Full} datasets are slightly unbalanced: 56.3\% of the records pertain to successes. For the \textit{Partial} datasets, the respective fraction is 51.17\%. \\
The data were randomly divided into training set (75\%) and test set (25\%). To avoid data leakage, therapies belonging to the same patient were included in the same set. A linear Support Vector Machine for classification was trained to predict success, or failure labels based on outputs representing probability estimates \citep{Platt1999}. 
A random search of the hyperparameters was carried out in order to determine the parameters for the best possible prediction performance. Values of hyperparameters were sampled from a probability distribution, and the performance of the resulting models was evaluated by a five times five-fold cross-validation. In this way, for each parameter setting, we obtain 25 performance values of the model with that setting when using different training and validation sets. In particular, the regularization parameter \textit{C} was sampled from a log-uniform distribution $\displaystyle( C\sim U(e^{-14},1))$. For each cross-validation set, the model with the lowest value of \textit{C} whose average performance (in terms of ROC-AUC score) was not significantly lower than the best average performance was selected (Benjamini-Hochberg-corrected Wilcoxon signed-rank test \citep{Wilcoxon1945IndividualCB} with a significance threshold of 0.05). For the mutations for which no data was available to fit the parameter $\theta$ of the sigmoid curve, $\theta$ was treated as a hyperparameter of the model, sampled uniformly between the minimum and the maximum value of the parameters $\alpha$ and $\beta$ learned for the mutations of the same drug class.

\subsection{Results}
Model performances were compared performing the Nadeau and Bengio statistical test \citep{NIPS1999_7d12b66d} on ROC-AUC scores, with a significance level of 5\%.\\ 

\textbf{\textit{Partial\_model} results.}
The results obtained from the two models based on the partial datasets are shown in Table \ref{tab:results partial dataset}. 
The \textit{Partial\_History\_Weighted model} achieves almost two percentage points more in ROC-AUC score than the \textit{Partial\_No-history\_Non-weighted model}, exhibiting a statistically significant difference between the two approaches. \\

\textbf{\textit{Full\_model} results.}
Table \ref{tab: results full dataset} shows the results obtained using the four models based on the full datasets. Thanks to the availability of more data points, the models obtain better results overall compared to the Partial models.
In particular, the \textit{Full\_History\_Weighted model} reaches the highest ROC-AUC score of 76.34\% (±0.099). The p-values associated with the ROC-AUC scores of the models that do not account for or partially account for historical information when compared with the \textit{Full\_History\_Weighted model} show that the difference in performance is statistically significant. \\

\begin{table}[!t]
\centering
\scalebox{0.7}{
\renewcommand{\arraystretch}{1.1}
\begin{tabular}{lllll}
\toprule
Model & AUC & Acc & Rec & Spec  \\
\midrule
\textit{Partial\_History\_Weighted} & \begin{tabular}[c]{@{}l@{}} \textbf{72.42} \\ \small(±0.140) \end{tabular} & \begin{tabular}[c]{@{}l@{}}  67.45 \\ \small(±0.133) \end{tabular} & \begin{tabular}[c]{@{}l@{}} 61.85 \\ \small(±0.196) \end{tabular} & \begin{tabular}[c]{@{}l@{}}  72.17 \\ \small(±0.166) \end{tabular}\\
\arrayrulecolor[HTML]{D9D9D9}\hline
\textit{Partial\_No-history\_Non-weighted} & \begin{tabular}[c]{@{}l@{}}70.43\\ \small(±0.139) \\ \tiny{*}0.0011\end{tabular} & \begin{tabular}[c]{@{}l@{}} 65.33   \\ \small(±0.129)  \end{tabular}& \begin{tabular}[c]{@{}l@{}}  62.84 \\ \small(±0.204) \end{tabular}& \begin{tabular}[c]{@{}l@{}} 67.43 \\ \small(±0.165) \end{tabular}\\
\bottomrule
\end{tabular}}
\begin{tablenotes}
\small
\item[$^{*}$] indicates the p-value w.r.t. the history\_weighted model
\end{tablenotes}
\caption{Performance metrics of models trained with the Partial dataset. The metrics reported are ROC AUC score (AUC), Accuracy (Acc), Recall (Rec) and Specificity (Spec) in percentage (\%).}
\label{tab:results partial dataset}
\end{table}

\begin{table}[!t]
\centering
\scalebox{0.72}{
\renewcommand{\arraystretch}{1.1}
\begin{tabular}{lllll}
\toprule
Model & AUC & Acc & Rec & Spec  \\
\midrule
\textit{Full\_History\_Weighted} & \begin{tabular}[c]{@{}l@{}} \textbf{76.34} \\ \small(±0.099) \end{tabular} & \begin{tabular}[c]{@{}l@{}}  70.74 \\ \small(±0.087) \end{tabular} & \begin{tabular}[c]{@{}l@{}} 64.95 \\ \small(±0.148) \end{tabular} & \begin{tabular}[c]{@{}l@{}}  73.28 \\ \small(±0.010) \end{tabular}\\
\arrayrulecolor[HTML]{D9D9D9}\hline
\textit{Full\_No-history\_Weighted} & \begin{tabular}[c]{@{}l@{}}\textbf{76.13}\\ \small(±0.099) \\ \tiny{*}0.0064\end{tabular} &  \begin{tabular}[c]{@{}l@{}} 70.60 \\ \small(±0.088) \end{tabular} &   \begin{tabular}[c]{@{}l@{}} 65.31 \\ \small(±0.151)\end{tabular}&  \begin{tabular}[c]{@{}l@{}}  72.80 \\ \small(±0.101) \end{tabular}\\
\arrayrulecolor[HTML]{D9D9D9}\hline
\textit{Full\_History\_Non-weighted} & \begin{tabular}[c]{@{}l@{}}76.67\\  \small(±0.100) \\ \tiny{*}1.02e-32 \end{tabular} & \begin{tabular}[c]{@{}l@{}} 70.10 \\ \small(±0.090) \end{tabular} & \begin{tabular}[c]{@{}l@{}} 58.32  \\ \small(±0.147) \end{tabular}& \begin{tabular}[c]{@{}l@{}} 76.58 \\ \small(±0.101) \end{tabular}\\
\arrayrulecolor[HTML]{D9D9D9}\hline
\textit{Full\_No-history\_Non-weighted} & \begin{tabular}[c]{@{}l@{}}74.98\\ \small(±0.098) \\ \tiny{*}6.92e-23\end{tabular} & \begin{tabular}[c]{@{}l@{}} 69.60 \\ \small(±0.088)  \end{tabular}& \begin{tabular}[c]{@{}l@{}} 58.41 \\ \small(±0.144) \end{tabular}& \begin{tabular}[c]{@{}l@{}} 76.01 \\ \small(±0.102) \end{tabular}\\
\bottomrule
\end{tabular}}
\begin{tablenotes}
\small
\item[$^{*}$] indicates the p-value w.r.t. the history\_weighted model
\end{tablenotes}
\caption{Performance metrics of models trained with the Full dataset. The metrics reported are ROC AUC score (AUC), Accuracy (Acc), Recall (Rec) and Specificity (Spec) in percentage (\%).}
\label{tab: results full dataset}
\end{table}

\begin{figure}[!t]
\centering
\begin{subfigure}[b]{0.32\textwidth}
    \includegraphics[width=\textwidth, height=4cm]{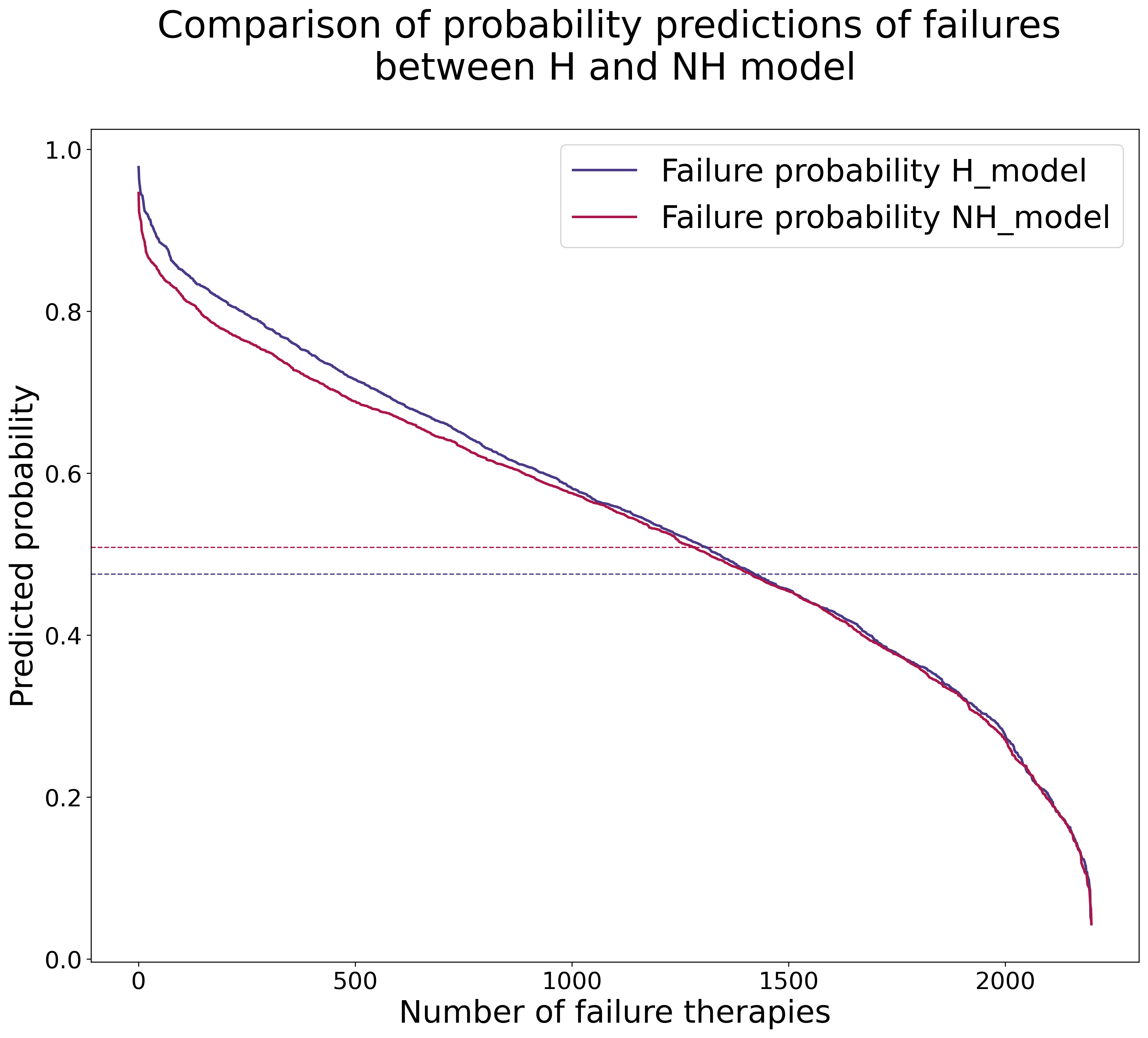}
    \caption{}
    \label{fig:sub2}
  \end{subfigure}
  \hfill
  \centering
\begin{subfigure}[b]{0.32\textwidth}
    \includegraphics[width=\textwidth, height=4cm]{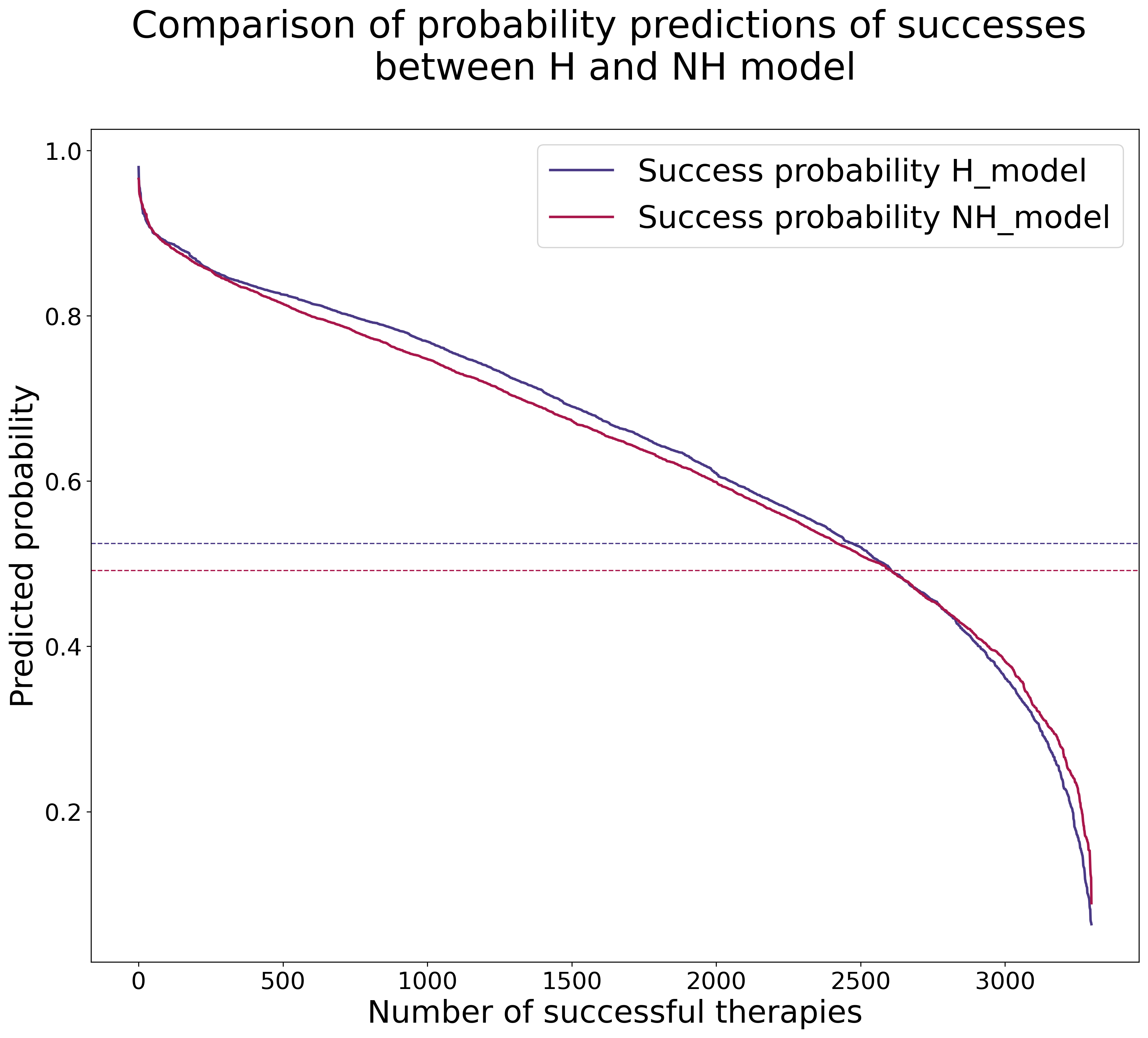}
    \caption{}
    \label{fig:sub4}
  \end{subfigure}
  \caption{
  Plots of the probability distributions predicted by the models, respectively (a) for successes and (b) for failures. On the x-axis, there is the number of therapies, and on the y-axis, the models' predicted probabilities. The blue and red dotted lines represent the cut-offs for probabilities, respectively, for the H and NH-model. For each model, the portion of the line above its cut-off represents therapy correctly classified by the model, while the one below represents therapy incorrectly classified by the model.}
  \label{fig:grid}
\end{figure}

Additional statistical analysis was conducted to assess the impact of incorporating historical information. Specifically, an attempt was made to compare the performance of the \textit{Full\_history\_weighted} (H) model and the \textit{Full\_No-history\_Non-weighted} (NH) model. This comparison aimed to evaluate the influence of historical information on the predictive accuracy of the models.
Table \ref{tab:predicted_probabilities} presents the mean and standard deviation of predicted probabilities for treatment successes and failures, ranked according to whether or not historical information was considered. Notably, when only treatments with a history (indicated with a $\checkmark$) are considered, i.e. type-1 therapy-patient pairs, the H-model consistently outperforms the NH-model in predicting the probabilities of both success and failure. This suggests that incorporating historical information provides valuable insights for accurate prediction.\\
Wilcoxon tests were performed on the probability distributions of treatment outcomes between the H-model and NH-model to assess the statistical differences between the probabilities predicted by the two models.
The respective null hypotheses are given in Table \ref{tab: wilcoxon_test}. Results show significant differences between the H and NH models. For both failures, with and without historical information, the H-model outperforms the NH-model, as evidenced by the remarkably low p-values ($5.32e^{-16}$ and $9.03e^{-25}$, respectively). Regarding all treatment successes, the H-model significantly outperforms the NH-model, with a p-value of $7.15e^{-6}$. However, when narrowing down the analysis to only successes with historical information, the high p-value ($0.9536$) indicates that the null hypothesis that the H-model predicts smaller or equal probabilities than the NH-model cannot be rejected. This can be interpreted to indicate that historical information does not play a significant role in the prediction of successful therapies.\\ 
Figure \ref{fig:grid} displays the H and NH-model's distributions of predicted probabilities for successes and failures. For both models, the cut-offs for probabilities are represented for both models by lines parallel to the x-axis. For ease of identification, each cut-off line is colored to match the color of its associated model. These cut-offs for probabilities are the threshold values that determine the class assignment for each data point based on the probabilities predicted by the models, and they have been tuned as reported in the Supplementary material. For each model in the figure, the part of the line above its cut-off represents successful or failing therapies correctly classified by the model, while the portion below shows therapies incorrectly classified. In general, the H-model seems to predict higher probabilities than the NH-model. Focusing on failures, Figure \ref{fig:sub2} shows that the H-model correctly classifies more failures than the NH-model. Regarding the space between the two cut-offs, where failures are correctly classified by the H-model but not by the NH-model, the H's probabilities are slightly higher. Curve segments below the lower cut-off represent failures misclassified by both models, with comparable probability distributions. Focusing on successes, Figure \ref{fig:sub4} shows fewer successes classified correctly by the H-model than the NH-model. However, the probabilities predicted by the H-model are higher, even for successes only correctly classified by the NH-model. NH-model's probabilities are higher for misclassified successes represented by lines below the lower cut-off. These data support the intuition that mutation profiling for each patient is useful for considering past resistance that may still play a role when changing therapy is necessary, to avoid failure. 

\begin{table}[!t]
\centering
\renewcommand{\arraystretch}{1.}

\begin{tabular*}{\columnwidth}{@{\extracolsep{\fill}}p{1.5cm}p{1.5cm}p{1.5cm}p{2cm}@{\extracolsep{\fill}}}
\toprule
   Type\textcolor{white}{\_}of therapies correctly classified & Type\textcolor{white}{\_}of model & Only therapy with history & Mean±SD\textcolor{white}{\_}of predicted probabilities \\
\midrule
        Successes & H & $\checkmark$ & 0.720 ± 0.11 \\
        \arrayrulecolor[HTML]{D9D9D9}\hline
        Successes & NH & $\checkmark$ & 0.699 ± 0.20 \\
        \arrayrulecolor[HTML]{D9D9D9}\hline
        Failures & H & $\checkmark$ & 0.699 ± 0.12 \\
        \arrayrulecolor[HTML]{D9D9D9}\hline
        Failures & NH & $\checkmark$ & 0.677 ± 0.099 \\
        \arrayrulecolor[HTML]{D9D9D9}\hline
        Successes & H & \textbf{$\times$} & 0.724 ± 0.106 \\
        \arrayrulecolor[HTML]{D9D9D9}\hline
        Successes & NH & \textbf{$\times$} & 0.700 ± 0.1146 \\
        \arrayrulecolor[HTML]{D9D9D9}\hline
        Failures & H & \textbf{$\times$} & 0.667 ± 0.119 \\
        \arrayrulecolor[HTML]{D9D9D9}\hline
        Failures & NH & \textbf{$\times$} & 0.665 ± 0.098 \\
   \botrule
\end{tabular*}
   \begin{tablenotes}
\small
\item[$^{*}$] indicates the p-value w.r.t. the history\_weighted model
\end{tablenotes}
    \caption{Mean and standard deviation of predicted probabilities by the two models, divided by type of therapy (success or failure). The column "only therapy with history" is valued with $\checkmark$ when only therapies with more than one previous genotype are considered or with \textbf{$\times$} when all the therapies are considered.}
    \label{tab:predicted_probabilities}
\end{table}

\begin{table}[!t]
\centering
\renewcommand{\arraystretch}{1.}

\begin{tabular*}{\columnwidth}{@{\extracolsep{\fill}}p{6cm}p{1.5cm}@{\extracolsep{\fill}}}
\toprule
    Wilcoxon test Between H and NH probability distribution for: & P-VALUE \\
\midrule
successes  \newline $H_0: p(Success_H)\le p(Success_{NH})$ & \textbf{ 7.15e$^{-6}$}\\
        \arrayrulecolor[HTML]{D9D9D9}\hline
        successes with history \newline $H_0: p(Success_H)\le p(Success_{NH})$ & $0.9536$ \\
        \arrayrulecolor[HTML]{D9D9D9}\hline
       failures \newline $H_0: p(Failure_H)\le p(Failure_{NH})$ & \textbf{5.32e$^{-16}$} \\
        \arrayrulecolor[HTML]{D9D9D9}\hline
        failures with history \newline $H_0: p(Failure_H)\le p(Failure_{NH})$ & \textbf{9.03e$^{-25}$} \\
   \botrule
\end{tabular*}
   \begin{tablenotes}
\small
\item[$^{*}$] indicates the p-value w.r.t. the history\_weighted model
\end{tablenotes}
\caption{p-values of the Wilcoxon tests performed on the predicted probability distributions of the models, with the null hypothesis $H_0$. }
\label{tab: wilcoxon_test}
\end{table}

\section{Discussion}

We have conducted several statistical analyses in order to assess the impact of incorporating historical information into the prediction of the efficacy of anti-HIV drug therapy. \\
Models employing \textit{Partial} datasets, including only patient-therapy pairs with relevant medical history before the target therapy (type-1 pairs), exhibit significant differences in prediction performance. These underscore the importance of incorporating a patient’s history into predictions of treatment outcomes, especially when eliminating the data dilution associated with therapies that lack the previous history.\\
Deeper analyses were conducted on the \textit{Full} models for the importance of training on \textit{Full} datasets. This includes both patient-therapy pairs with relevant medical history before the target therapy and patients without historical information. The latter include patients for which limited data are available or those on first-line therapy. Ensuring accurate predictions for both categories is essential for providing valuable insights for clinical decision-making. 
The \textit{Full\_History\_Weighted model} (H) model has a higher ROC-AUC score (76.34\%) than the \textit{Full\_No-history\_No-weighted model} (NH) (74.98\%). Interestingly, the \textit{Full\_No-history\_Weighted model} demonstrates performance comparable to the \textit{Full\_History\_Weighted model}. Both the models take historical information into account to some extent: the former because it considers the mutations of all the GRTs performed in the past and weighted as explained earlier; the latter because, although only the last GRT is considered, the mutations are weighted taking into account the elapsed time since the mutation was detected, that is the date on which the virus was sequenced. This suggests that the substantial improvements observed with the history-weighted model can be attributed mainly to incorporating the weighting factors rather than including historical mutations per se. 

One reason why the H-model has better predictive performance than the NH-model, especially in the case of failures, could be attributed to the possibility that historical mutations represent latent reservoirs for HIV.  Latent reservoirs are cells infected with the virus but in tissues other than blood serum. These remain inactive or dormant, evading the body's immune response to standard ART. The existence of latent reservoirs makes it difficult to eradicate HIV from the body because these cells can reactivate and generate new viral particles, leading to a reappearance of the virus and treatment failure. Presumably, the model, considering weighted historical mutations, can capture the intricate relationship between latent reservoirs and response to therapy, improving predictive performance in case of treatment failure.

Our study's implications align with a concept emerging from a recent study conducted by part of our team to study mutational history in a different context \citep{Pirkl_Lengauer_2023}. In this work, data from various patients are used cross-sectionally rather than longitudinally to infer the accumulation of mutations in multidrug-resistant patients, i.e., the order in which mutations occur over time (mutational history). Therefore, the respective model can predict aspects of mutational history from current genotypes. For example, if we observe a genotype with the X mutation and not with the Y mutation, but in the mutational history, Y was observed before X, Pirkl et al. \citep{Pirkl_Lengauer_2023} hypothesize that the Y mutation may now be present only in the latent reservoir genotypes and not in the blood serum and that mutations are still present in the latent reservoir and thus have an impact on future therapies. Both studies emphasize the importance of mutations that occurred in the past for future therapies and drug resistance, respectively.

Additional experiments with the prediction of treatment outcome have been carried out using the Stanford Treatment Change episode database available at \url{https://hivdb.stanford.edu/TCEs/}. The results are reported in the section \textit{Additional results} of the Supplementary material and confirm the insights on the importance of historical mutations.

In conclusion, our study sheds light on the fact that incorporating the temporal dynamics of virus acquiring mutations in response to drug therapy improves prediction accuracy compared with the standard analysis of only the last available genotype. The results underscore the importance of considering mutation dynamics and its potential influence on treatment outcomes. They provide valuable insights into the complex dynamics of HIV infection, guiding future research and informing the development of effective therapeutic strategies. \\
Limitations of our work include the potential suboptimality of the constructed weighting factor for mutations, which combines VL magnitude, mutation timing, and duration, without a clear understanding of the individual influences of these three aspects. Furthermore, although the differences in performance between the models that include history and those that do not are statistically significant, the ROC-AUC of the non-history models is relatively high. This could be due to the methodology used to include history or the inherent incompleteness of the available data.

Our findings can be interpreted in two ways. On the one hand, the fact that historical mutations impact therapy prediction in a statistically highly significant fashion points to the importance of involving mutational history in the research on therapy prediction. Of note, this implies continuing to consider past data when investigating response to novel and future treatments as a general research plan. On the other hand, the fact that the impact is quite small in absolute terms means that the clinical relevance of incorporating historical mutations in the analysis of viral resistance for an individual patient is debatable. This can be viewed as support for alleviating the clinician from the need to elaborate on and input complex data into a genotypic prediction system, although reasoning on past treatment and resistance data remains well established in the clinician's choice of a new therapy. 


\section{Competing interests}
No competing interest is declared.

\section{Author contributions statement}
G.D.T, M.P, R.K., L.P, M.Z., T.L. conceptualized the study. G.D.T.,F.I.,I.V, and A.S. managed data curation and access. G.D.T. processed the data and conducted the statistical analysis under the supervision of all the authors. G.D.T., M.P., R.K., M.Z., and T.L. interpreted the results. G.D.T. wrote the manuscript.  M.P., M.Z., L.P. and T.L. reviewed the manuscript. All authors have read and agreed to the published version of the manuscript.

\section{Acknowledgments}
The authors would like to thank the EuResist Network working group for their valuable work for the EIDB. 

\section{Data availability statement}
This analysis was conducted using the EIDB. For further validation, we encourage reproducing this study with the latest release of the EIDB, which can be accessed upon request through the Euresist Network at \url{https://www.euresist.org/become-a-partner}. Due to the sensitive nature of personal medical data, it is not feasible to make this data publicly available on the internet. \\ Additionally, data from the HIVdb were utilized in this study. The HIVdb is openly accessible at \url{https://hivdb.stanford.edu/TCEs/}.

\section{Ethic statement}
Ethical approval was granted in the host countries of the respective original databases contributing data to EIDB.

\bibliographystyle{cas-model2-names}

\bibliography{cas-refs}

\appendix
\section*{Supplemental material}

\section{Details on the datasets and on results}
The drugs considered in the datasets are the following: lamivudine (3TC), abacavir (ABC), amprenavir (APV), atazanavir (ATV), zidovudine (AZT), bictegravir (BIC), cabotegravir (CAB), stavudine (D4T), zalcitabine (DDC), didanosine (DDI), delavirdine (DLV), doravirine (DOR), darunavir (DRV), dolutegravir (DTG), efavirenz (EFV), etravirine (ETR), elvitegravir (EVG), fosamprenavir (FPV), emtricitabine (FTC), indinavir (IDV), lopinavir (LPV), nelfinavir (NFV), nevirapine (NVP), raltegravir (RAL), rilpivirine (RPV), saquinavir (SQV), tenofovir alafenamide (TAF), tenofovir disoproxil (TDF), tipranavir (TPV).

\begin{figure*}[!t]
  \centering
 \includegraphics[width=\textwidth]{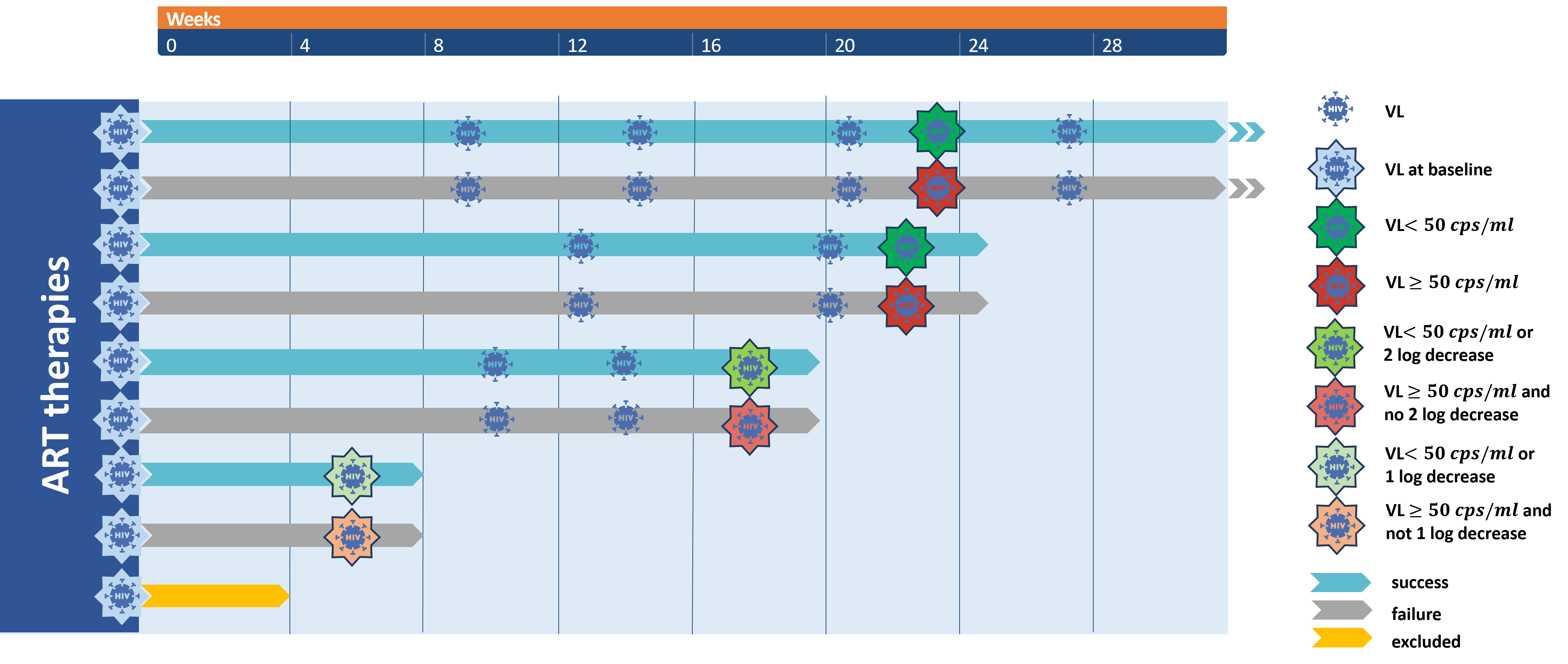}
 
\caption{Graphical representation of how to label an ART therapy as success or failure according to the Standard Datum definition.}
  \label{fig:standard_datum}
\end{figure*}

Due to the definitions of PTEs (Patient-Treatment Episodes) and PTCEs (Patient-Treatment Change Episodes), it is feasible for various entries in the datasets, involving distinct treatments, to relate to a single patient undergoing different therapies in different time periods.
Table \ref{tab:therapies per patient} reports how many therapies there are for the same patient, divided between successful (label 0) and failed therapies (label 1), according to the Standard Datum definition provided in the main text, in Section 3.2. A graphical representation of the Standard Datum is given in Figure \ref{fig:standard_datum}.  
To guard against data leakage, either all therapies referring to the same patient were included in the training set or all of them were included in the test set.

Four different datasets have been built and six different linear-SVM models have been trained. A tabular form of listing the various models and datasets used in a simple visualization is presented in Table \ref{tab:Datasets and models' characteristics}.

\begin{table}[ht]
\centering
\renewcommand{\arraystretch}{1.}

\begin{tabular*}{\columnwidth}{@{\extracolsep\fill}llll@{\extracolsep\fill}}
\toprule
    \# Therapies per patient & Label & \# Patients \\

\midrule
1	& 0	& 4659 \\
1	& 1	& 3363 \\
\arrayrulecolor[HTML]{D9D9D9}\hline
2	& 0	& 1781 \\
2	& 1	& 1230\\
\arrayrulecolor[HTML]{D9D9D9}\hline
3	& 0	& 691\\
3	& 1	& 501\\
\arrayrulecolor[HTML]{D9D9D9}\hline
4	& 0	& 253\\

4	& 1	& 235\\
\arrayrulecolor[HTML]{D9D9D9}\hline
5	& 0	& 96\\

5	& 1	& 118\\
\arrayrulecolor[HTML]{D9D9D9}\hline
6	& 0	& 44\\

6	& 1	& 49\\
\arrayrulecolor[HTML]{D9D9D9}\hline
7	& 0	& 19\\

7	& 1	& 26\\
\arrayrulecolor[HTML]{D9D9D9}\hline
8	& 0	& 14\\

8	& 1	& 14\\
\arrayrulecolor[HTML]{D9D9D9}\hline
9	& 0	& 4\\

9	& 1 & 	10\\
\arrayrulecolor[HTML]{D9D9D9}\hline
10	& 0	& 5\\
\arrayrulecolor[HTML]{D9D9D9}\hline
11	& 1	& 3\\

11	& 1	& 1\\
\arrayrulecolor[HTML]{D9D9D9}\hline
12	& 1	& 2\\
\arrayrulecolor[HTML]{D9D9D9}\hline
15	& 1	& 1\\

   \botrule
  \end{tabular*}
  \caption{Number of therapies per patient}
  \label{tab:therapies per patient}
\end{table}

\begin{table*}[t]
\renewcommand{\arraystretch}{1.1}
\begin{tabular*}{\textwidth}{@{\extracolsep{\fill}}p{3.5cm}p{2cm}p{2cm}p{2cm}p{2cm}p{2cm}p{1.8cm}@{\extracolsep{\fill}}}
\toprule
Model name & Possible \# of GRTs $>$ 1 prior to the target therapy & \# of GRTs = 1 prior to the target therapy & Mutations\textcolor{white}{\_}of the\textcolor{white}{\_}GRT\textcolor{white}{\_}at baseline & The mutations of GRTs before the\textcolor{white}{\_}one\textcolor{white}{\_}at baseline & Binary\textcolor{white}{\_}vector of mutations & Weighted vector\textcolor{white}{\_}of mutations\\
\midrule
\textit{Partial\_History\_Weighted} & $\checkmark $ & \textbf{$\times$} & $\checkmark$ & $\checkmark$ & $\checkmark$ & $\checkmark$\\
\arrayrulecolor[HTML]{D9D9D9}\hline
 \textit{Partial\_No-history\_Non-weighted} & $\checkmark$ & \textbf{$\times$} & $\checkmark$ & \textbf{$\times$} & $\checkmark$ & \textbf{$\times$}\\
\arrayrulecolor[HTML]{D9D9D9}\hline
\textit{Full\_History\_Weighted} & $\checkmark$ & $\checkmark$ &$\checkmark$ & $\checkmark$ & $\checkmark$ & $\checkmark$ \\
\arrayrulecolor[HTML]{D9D9D9}\hline
 \textit{Full\_No-history\_Weighted} & $\checkmark$ & $\checkmark$ & $\checkmark$ & \textbf{$\times$} & $\checkmark$ & $\checkmark$ \\
 \arrayrulecolor[HTML]{D9D9D9}\hline
 \textit{Full\_History\_Non-weighted} & $\checkmark$ & $\checkmark$ & $\checkmark$ & $\checkmark$ & $\checkmark$ & \textbf{$\times$}\\
\arrayrulecolor[HTML]{D9D9D9}\hline
 \textit{Full\_No-history\_Non-weighted} & $\checkmark$ & $\checkmark$ & $\checkmark$ & \textbf{$\times$} & $\checkmark$ & \textbf{$\times$}\\
\botrule
\end{tabular*}
\caption{Datasets and models' characteristics}
  \label{tab:Datasets and models' characteristics}
\end{table*}

\subsection{Significance test}
To assess whether the history model performs better than the no-history model in terms of generalization capability, we use significance tests. The most common method for comparing the performance of Machine Learning models is the paired Student's t-test combined with random subsampling of the training set. However, one of the key assumptions of this test is that the underlying data be sampled independently from the two populations being compared. Since the models are trained on the same training set, the paired Student's t-test could lead to misleading results, with a high false positive rate (i.e., having a high probability of rejecting the null hypothesis indicating that the two models are significantly different, when this is in fact the case, i.e. overrating the significance of using history data). Nadeau and Bengio \citep{NIPS1999_7d12b66d} propose a variance correction that accounts for the dependence between the two. We apply their significance test in our analysis, with a significance level of 5\% .

\subsection{Cut-off for the predicted probabilities}
The trained models generate probabilities that indicate the likelihood of belonging to a certain class. Typically, a default threshold of 0.5 is used to assign class labels based on these probabilities. However, relying solely on this default threshold can lead to suboptimal performance, particularly when dealing with datasets that have unbalanced class distributions. Although, in our case, the overall dataset is not highly unbalanced, it is still helpful to adjust the threshold according to the characteristics of the training data.\\
A simple strategy was used to determine the most suitable threshold for mapping probabilities onto class labels. A comprehensive search was conducted with 1,000 threshold values equally spaced between 0 and 1. Through this iterative process, each model identified an individual threshold that maximized balanced accuracy. This ensures that the threshold is tailored to the specific characteristics of each model, finding a desirable compromise between sensitivity and specificity.\\
Through this iterative threshold selection approach, we improve model performance and obtain more accurate classification results. It allows us to effectively handle variations within the dataset and address situations where imbalances between classes may affect the model's predictive capabilities. 

\section{Analysis of the importance of mutations}
\subsection{Mutations ranking} \label{sec:mut_ranking}

Identifying the role of mutations in predicting the outcome of therapy is paramount for various reasons. The response of HIV to specific antiretroviral drugs can vary depending on the mutations present in its genome. Gaining an understanding of the role of mutations is crucial for improving treatment recommendations and maximizing treatment effectiveness. It enables us to identify mutations associated with drug resistance and to comprehend how these mutations interact and impact viral suppression. This knowledge aids in selecting appropriate drug regimens that optimize treatment outcomes.

The model trained is a linear Support Vector Machine. The linear SVM provides hyperplane coefficients that represent the weights assigned to the features of the input data. These coefficients serve for assessing the importance and contributions of each feature in the SVM's decision-making process. The higher the absolute value of a mutation's coefficient, the more that mutation plays a role in determining the outcome. Moreover, we calculated individual weights for each mutation of each patient, as described in the main paper in Section 2.2. 
If a mutation is assigned a low weight, at least one of the following conditions occurs: \textit{(i)} it was observed far back in time, \textit{(ii)} the patient's viral load, at the time of observing the mutation, was a low level or undetectable, \textit{(iii)} the Stanford score associated with the mutation is low. The low weight associated with a mutation suggests that the mutation does not highly impact therapy outcome. 
\\
In studying the impact of mutations, our objectives were twofold:
\begin{itemize}
    \item To identify the mutations that consistently have a significant impact on therapy outcome.

    \item To identify specific mutations that are assigned a high model coefficient despite entering the model with a low weight.
\end{itemize}
For the first objective, we simply ranked of mutations according to the absolute values of their SVM coefficients. Inspecting Table \ref{tab: coefficient_ranking} we can make interesting observations. For example, the top-ranking mutation is T200K. This mutation is not widely recognized as an important mutation for drug resistance in HIV but a few studies reported that T200K, even though it had not been reported previously, might be associated with NVP resistance \citep{Hachiya2004-wa}. In general, many of the top-ranked mutations in Table \ref{tab: coefficient_ranking} are not recognized resistance mutations. Their coefficient values may be high because they were frequently observed near the start of the therapy of interest (i.e., the most recent genotype) and/or in the presence of a high viral load. Whereas most of the literature focuses on mutations that reduce viral fitness as something potentially exploitable in the clinic, these mutations could instead increase viral fitness and, thus, viral load.

\begin{figure}[!t]
  \centering
  \begin{subfigure}[b]{\columnwidth}
    \includegraphics[width=.90\columnwidth, height=5.5cm]{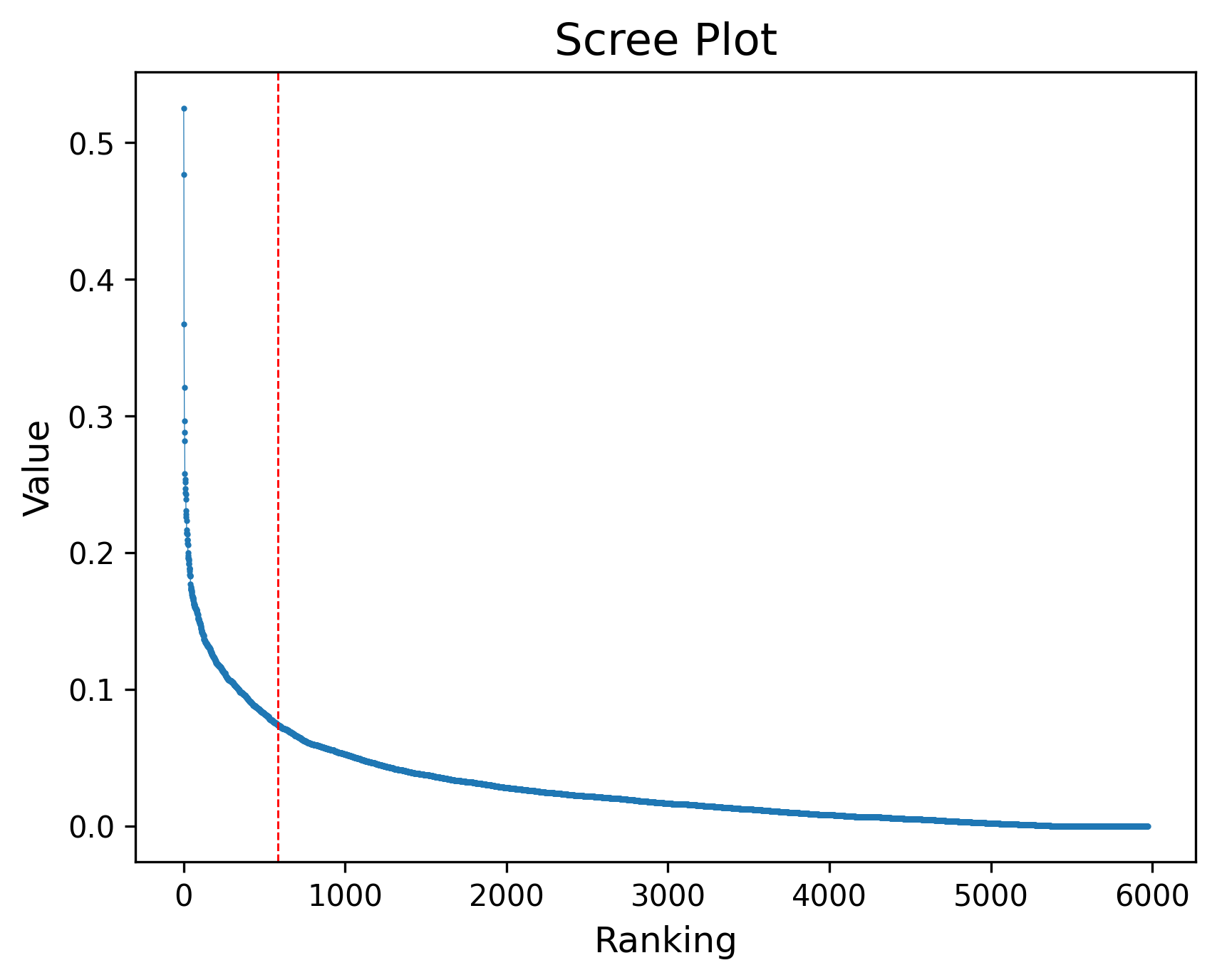}
    \caption{}
    \label{fig:sub1_}
  \end{subfigure}
  \hfill
\begin{subfigure}[b]{\columnwidth}
    \includegraphics[width=0.90\textwidth, height=5.4cm]{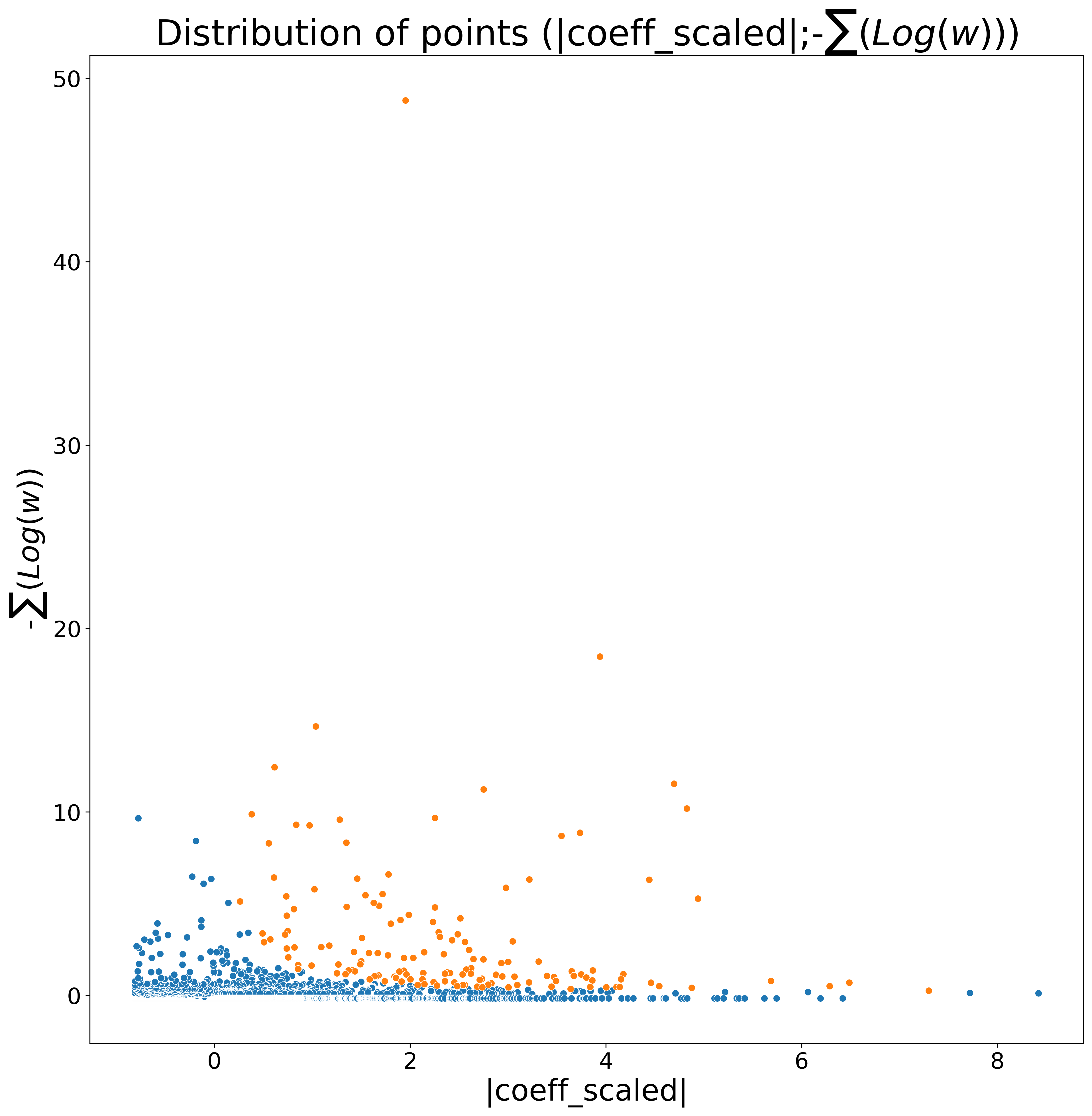}
    \caption{}
    \label{fig:sub2_}
  \end{subfigure}
 
\caption{Figure (a) is the scree plot of the mutation, based on the ranking values obtained by equation \ref{eq_ranking}. Figure (b) is the scatter plot of the points ( $|coefficient_m|_{z\_scaled}$;$-\sum_{i=1}^{n}{\log{w_m^i}}$) to show that the ranking does not depend only to one of the two factors of the ranking formula but by both of them. The orange points are the mutations identified in the scree plot before the dashed red line that is the flexion point of the curve.}
  \label{fig:plots_ranking_coef}
\end{figure}

For the second objective, a ranking formula was created to consider the potential discrepancy between the SVM coefficient and the customized weights, allowing mutations with high coefficients but low weights to still receive a significant ranking value. The ranking formula for each mutation $m$ is the following: 
\begin{equation}\label{eq_ranking}
\text{Ranking\_value}_m = \lvert\text{coefficient}_m\rvert_{\text{z\_scaled}} \cdot (-\sum_{i=1}^{n} \log{w_m^i})_{\text{z\_scaled}}
\end{equation}
where \textit{coefficient} is the SVM hyperplane coefficient, $n$ is the size of the training set and $w_m^i$ is the weight learned for mutation $m$ of therapy-patient tuple $i$. We obtained the positional ranking of the mutations by decreasingly sorting the absolute \textit{$Ranking\_value$} for each mutation.
From this ranking, we identify 149 mutations that could have the impact we aim for, which are
the ones on the left of the dotted red line, representing the flexion point of the curve, in Figure \ref{fig:sub1_}. That is, we applied the elbow method on the scree plot in the figure. The top-ranking 30 mutations are listed in Table \ref{tab: ranking_lowWeight_highCoef}. Some observations considering this ranking may be clinically relevant. For example, L63P, among the top-ranked mutations, is usually not considered a resistance mutation. However, it has been pointed out that \textit{(i)} this mutation may persist for even more than 18 months despite therapy changes \citep{Pao2004-ab}, \textit{(ii)} the mutation by itself does not render the virus resistant to inhibitors, but it does help the virus replicate more effectively, especially when under pressure from drugs. This suggests that small changes in the virus can have significant compensating effects, which could contribute to the evolution of drug-resistant variants of HIV-1 \citep{Sune2004-bi}. Indeed, most of mutations included in Table \ref{tab: ranking_lowWeight_highCoef} that are not considered resistance mutations have been implicated in compensatory or resistance modulation mechanisms. In PR, L10I, L63P, I93L even had a Stanford HIVdb score (although low) in past versions. The same applies to RTE44D and RTH208Y which have been studied in detail \citep{Betancor2014-tk,Girouard2003-ex,Nebbia2007-jd,Romano2002-ny}.
Being aware that mutations that are not considered major could increase viral fitness and have a compensatory effect that helps the virus replicate more effectively while not making the virus resistant to inhibitors per se, could have important implications for clinical practice. In particular, rule-based genotypic interpretation systems that consider only major resistance mutations may not be sufficient to design a therapy that considers all factors that play a role in the long-term efficacy of therapy.

 \begin{table}[!t]
\centering
\renewcommand{\arraystretch}{1.}

\begin{tabular*}{\columnwidth}{@{\extracolsep\fill}llll@{\extracolsep\fill}}
\toprule
    Mutation & Value coefficient H model \\

\midrule
 RTT200K & 0.3207 \\ \arrayrulecolor[HTML]{D9D9D9}\hline
        RTD113N & 0.2964 \\ \arrayrulecolor[HTML]{D9D9D9}\hline
        INK14R & 0.2819 \\ \arrayrulecolor[HTML]{D9D9D9}\hline
        RTT200V & 0.2536 \\ \arrayrulecolor[HTML]{D9D9D9}\hline
        RTS322A & -0.2513 \\ \arrayrulecolor[HTML]{D9D9D9}\hline
        INM154I & -0.2466 \\ \arrayrulecolor[HTML]{D9D9D9}\hline
        RTK249R & -0.2434 \\ \arrayrulecolor[HTML]{D9D9D9}\hline
        RTK122Q & -0.2390 \\ \arrayrulecolor[HTML]{D9D9D9}\hline
        PRL63C & 0.2278 \\ \arrayrulecolor[HTML]{D9D9D9}\hline
        RTM16V & -0.2258 \\ \arrayrulecolor[HTML]{D9D9D9}\hline
        RTE79K & -0.2235 \\ \arrayrulecolor[HTML]{D9D9D9}\hline
        \textbf{PRV82S} & 0.2165 \\ \arrayrulecolor[HTML]{D9D9D9}\hline
        RTV245I & -0.2145 \\ \arrayrulecolor[HTML]{D9D9D9}\hline
        RTQ145E & 0.2136 \\ \arrayrulecolor[HTML]{D9D9D9}\hline
        RTQ207H & -0.2095 \\ \arrayrulecolor[HTML]{D9D9D9}\hline
        PRG17D & 0.2090 \\ \arrayrulecolor[HTML]{D9D9D9}\hline
        PRH69Y & 0.2068 \\ \arrayrulecolor[HTML]{D9D9D9}\hline
        RTP176S & -0.2058 \\ \arrayrulecolor[HTML]{D9D9D9}\hline
        PRI93L & -0.1999 \\ \arrayrulecolor[HTML]{D9D9D9}\hline
        \textbf{RTF227L} & 0.1977 \\ \arrayrulecolor[HTML]{D9D9D9}\hline
        RTS48P & 0.1960 \\ \arrayrulecolor[HTML]{D9D9D9}\hline
        \textbf{RTT215Y} & 0.1959 \\ \arrayrulecolor[HTML]{D9D9D9}\hline
        RTY271F & 0.1950 \\ \arrayrulecolor[HTML]{D9D9D9}\hline
        RTE194D & -0.1939 \\ \arrayrulecolor[HTML]{D9D9D9}\hline
        RTV35E & 0.1919 \\ \arrayrulecolor[HTML]{D9D9D9}\hline
        \textbf{PRL90M} & 0.1914 \\ \arrayrulecolor[HTML]{D9D9D9}\hline
        RTL210S & 0.1885 \\ \arrayrulecolor[HTML]{D9D9D9}\hline
        RTE248N & -0.1882 \\ \arrayrulecolor[HTML]{D9D9D9}\hline
        PRL19T & 0.1878 \\ \arrayrulecolor[HTML]{D9D9D9}\hline
        INV37I & 0.1862 \\
   \botrule
  \end{tabular*}
\caption{First 30 mutations of the ranking obtained ordering the mutations with descending absolute value of H model coefficients. The mutations name is preceded by PR (protease), RT (reverse transcriptase) or IN (integrase), depending on the region of the HIV genome. In bold, Stanford mutations.}
\label{tab: coefficient_ranking}
\end{table}

 \begin{table}[!t]
\centering
\renewcommand{\arraystretch}{1.}

\begin{tabular*}{\columnwidth}{@{\extracolsep\fill}llll@{\extracolsep\fill}}
\toprule
    Mutation & Ranking value \\

\midrule
        
        \textbf{RTM184V} & 95.30 \\ \arrayrulecolor[HTML]{D9D9D9}\hline
        \textbf{RTK103N} & 72.80 \\ \arrayrulecolor[HTML]{D9D9D9}\hline
        \textbf{PRL90M} & 54.19 \\ \arrayrulecolor[HTML]{D9D9D9}\hline
        \textbf{RTT215Y} & 49.21 \\ \arrayrulecolor[HTML]{D9D9D9}\hline
        \textbf{RTK65R} & 33.17 \\ \arrayrulecolor[HTML]{D9D9D9}\hline
        RTD123E & 30.88 \\ \arrayrulecolor[HTML]{D9D9D9}\hline
        \textbf{RTK70R} & 30.86 \\ \arrayrulecolor[HTML]{D9D9D9}\hline
        PRL63P & 28.02 \\ \arrayrulecolor[HTML]{D9D9D9}\hline
        PRI93L & 26.13 \\ \arrayrulecolor[HTML]{D9D9D9}\hline
        RTK122E & 21.81 \\ \arrayrulecolor[HTML]{D9D9D9}\hline
        \textbf{RTM184I} & 20.33 \\ \arrayrulecolor[HTML]{D9D9D9}\hline
        \textbf{RTK219Q} & 17.46 \\ \arrayrulecolor[HTML]{D9D9D9}\hline
        \textbf{PRI84V} & 15.21 \\ \arrayrulecolor[HTML]{D9D9D9}\hline
        \textbf{RTD67N} & 12.26 \\ \arrayrulecolor[HTML]{D9D9D9}\hline
        RTK277R & 11.74 \\ \arrayrulecolor[HTML]{D9D9D9}\hline
        RTG196E & 11.23 \\ \arrayrulecolor[HTML]{D9D9D9}\hline
        RTE44D & 10.82 \\ \arrayrulecolor[HTML]{D9D9D9}\hline
        RTK49R & 10.58 \\ \arrayrulecolor[HTML]{D9D9D9}\hline
        \textbf{RTT215F} & 9.50 \\ \arrayrulecolor[HTML]{D9D9D9}\hline
        \textbf{RTL210W} & 9.28 \\ \arrayrulecolor[HTML]{D9D9D9}\hline
        PRL10I & 9.00 \\ \arrayrulecolor[HTML]{D9D9D9}\hline
        RTP294T & 8.99 \\ \arrayrulecolor[HTML]{D9D9D9}\hline
        \textbf{PRV82A} & 8.94 \\ \arrayrulecolor[HTML]{D9D9D9}\hline
        \textbf{RTL100I} & 8.74 \\ \arrayrulecolor[HTML]{D9D9D9}\hline
        RTI202V & 8.43 \\ \arrayrulecolor[HTML]{D9D9D9}\hline
        RTS162C & 8.32 \\ \arrayrulecolor[HTML]{D9D9D9}\hline
        RTI293V & 8.24 \\ \arrayrulecolor[HTML]{D9D9D9}\hline
        RTT286A & 8.20 \\ \arrayrulecolor[HTML]{D9D9D9}\hline
        RTH208Y & 7.88 \\ \arrayrulecolor[HTML]{D9D9D9}\hline
        PRI64V & 7.81 \\
   \botrule
  \end{tabular*}
\caption{First 30 mutations of the ranking obtained ordering the mutations with descending ranking value computed as in \ref{eq_ranking}. The mutations name is preceded by PR (protease), RT (reverse transcriptase) or IN (integrase), depending on the region of the HIV genome. In bold, Stanford mutations.}
\label{tab: ranking_lowWeight_highCoef}
\end{table}

\subsection{Enrichment analysis}
An enrichment analysis of the mutations was performed. Enriched mutations are those that have become more widespread or common within a particular population or sample compared to their initial occurrence. The implication of enriched mutations could be \textit{(i)} drug resistance; \textit{(ii)} disease progression: independent of drug resistance, such as through tropism change or increased virulence. \\
To perform this analysis, Stanford tables on \href{https://hivdb.stanford.edu/cgi-bin/MutPrevBySubtypeRx.cgi}{https://hivdb. stanford.edu/cgi-bin/MutPrevBySubtypeRx.cgi} were used to analyze whether the frequency of each mutation of interest increased from naïve patients to treatment-experienced patients. Specifically, at the linked website, each table presents the results of an enrichment analysis on the genotype sequences of patients treated with PI, or NRTI, or NNRTI as first-line therapy. Columns specify HIV subtypes (A, B, C, D, F, G, AE, AG), while rows detail mutations by amino acid deviating from consensus B. We focused our analysis on the HIV B subtype column, as it has the largest number of sequences available, allowing for a more robust and representative mutation enrichment analysis. The intersection between a mutation row and a subtype column highlights the frequency of patients with that mutation for the specified subtype. Where the frequency is greater than 1\% indicates an enriched mutation that therefore could be clinically or biologically significant.
Of the 149 mutations identified in Section \ref{sec:mut_ranking}, we counted how many are present in the tables just described. It resulted that 62\% of the 149 selected mutations become enriched. This confirms that the approach we used in this work in treatment-experienced patients captured mostly biologically meaningful mutations, either conferring or modulating resistance. Importantly, one-third of the scored mutations do not appear to be enriched following treatment but may play a role for the clinical outcome. Mutations or polymorphisms of this kind may be associated with functional constraints limiting their selection under commonly used therapies yet play a role in response to treatment because of effects that are independent of drug resistance, e.g., impact on innate or adaptive immunity and fitness effects.

\section{Additional results}
In this section, we present additional computational results obtained when using the Treatment Change Episode (TCE) repository available from the Stanford University HIV drug resistance database (\url{https://hivdb.stanford.edu/TCEs/}).
The TCE Repository presents over 1,500 cases of TCE. Each TCE is coded through an XML schema that includes key clinical information including \textit{(i)} previously administered ARTs; \textit{(ii)} plasma HIV-1 RNA load, CD4 cell counts, and results of genotypic resistance assessments at baseline; \textit{(iii)} administration of subsequent salvage therapies; and \textit{(iv)} measurements of plasma HIV-1 RNA levels made during the course of salvage therapy. \\
A key pre-processing step was implemented to harmonize the Stanford TCE data with those from the EIDB, for both structure and information content, ensuring consistency and compatibility within the experimental analysis. This procedure involved a selection of PTEs that \textit{(i)} met the specific criteria for evaluating therapeutic outcomes, as outlined in the Standard Datum definition within Section 2 of the main paper, and \textit{(ii)} had enough information to determine mutation weights, as explained in Section 3.2 in the main paper. This resulted in the identification of a restricted dataset of 562 PTEs. \\
In this dataset, only 63 PTEs (11.2\%) are labeled as successes. 
There are two main reasons why this dataset is heavily skewed toward treatment failures. First, data collection may not have placed emphasis on creating a representative sample of HIV treatments. When asked to contribute to the TCE repository clinicians may simply have provided available genetic sequences, leading to an over-representation of genotypic tests done in conjunction with treatment failures because this has been the most typical use of genotyping for many years, particularly when treatment failure was common. Second, due to timing of the request to contribute to the TCE repository, the data refers to ART administered before 2011, i.e. before widespread use of the most effective treatment regimens, including second-generation INSTI therapies. This resulted in the collection of ARTs with significantly higher failure rates than more recent treatments, further exacerbating the dataset’s skew toward therapy failures.

\begin{table}[!t]
\centering
\scalebox{0.74}{
\renewcommand{\arraystretch}{1.1}
\begin{tabular}{lllll}
\toprule
Model & AUC & Acc & Rec & Spec  \\
\midrule
\textit{Full\_History\_Weighted} &\begin{tabular}[c]{@{}l@{}} 66.1  \\ \small(±0.54) \end{tabular} & \begin{tabular}[c]{@{}l@{}}89.8 \\ \small(±0.16) \end{tabular} & \begin{tabular}[c]{@{}l@{}}94.2 \\ \small(±0.16) \end{tabular}& \begin{tabular}[c]{@{}l@{}}55.5 \\ \small(±0.65) \end{tabular}\\
\arrayrulecolor[HTML]{D9D9D9}\hline
\textit{Full\_No-history\_Weighted} & \begin{tabular}[c]{@{}l@{}}58.7\\ \small(±0.52) \\ \tiny{*}7.32e-62\end{tabular} &\begin{tabular}[c]{@{}l@{}} 79.0 \\ \small(±0.23) \end{tabular}&\begin{tabular}[c]{@{}l@{}} 85.2 \\ \small(±0.25) \end{tabular}& \begin{tabular}[c]{@{}l@{}}50.8 \\ \small(±0.60) \end{tabular}\\
\arrayrulecolor[HTML]{D9D9D9}\hline
\textit{Full\_History\_Non-weighted} & \begin{tabular}[c]{@{}l@{}}64.0\\ \small(±0.54) \\ \tiny{*}3.94e-7\end{tabular} &\begin{tabular}[c]{@{}l@{}} 77.2 \\ \small(±0.22) \end{tabular}& \begin{tabular}[c]{@{}l@{}}80.0 \\ \small(±0.24) \end{tabular}& \begin{tabular}[c]{@{}l@{}}55.5 \\ \small(±0.74) \end{tabular}\\
\arrayrulecolor[HTML]{D9D9D9}\hline
\textit{Full\_No-history\_Non-weighted} & \begin{tabular}[c]{@{}l@{}}57.7\\ \small(±0.53) \\ \tiny{*}2.45e-73\end{tabular} & \begin{tabular}[c]{@{}l@{}}73.7 \\ \small(±0.23) \end{tabular}& \begin{tabular}[c]{@{}l@{}}77.0\\ \small(±0.25) \end{tabular} &\begin{tabular}[c]{@{}l@{}} 47.6\\ \small(±0.68) \end{tabular}\\
\bottomrule
\end{tabular}}
 \begin{tablenotes}
\small
\item[$^{*}$] indicates the p-value w.r.t. the history\_weighted model
\end{tablenotes}
\caption{Performance metrics of models on the external validation set derived from Stanford TCE dataset, when models are trained on the EIDB. The metrics reported are ROC AUC score (AUC), Accuracy (Acc), Recall (Rec) and Specificity (Spec) in percentage (\%).}
\label{tab:res_ext_val}
\end{table}

\begin{table}[!t]
\centering
\scalebox{0.74}{
\renewcommand{\arraystretch}{1.1}
\begin{tabular}{lllll}
\toprule
Model & AUC & Acc & Rec & Spec  \\
\midrule
\textit{Full\_History\_Weighted} & \begin{tabular}[c]{@{}l@{}} 76.2 \\ \small(±1.10) \end{tabular} & \begin{tabular}[c]{@{}l@{}}  77.1 \\ \small(±0.49) \end{tabular} & \begin{tabular}[c]{@{}l@{}} 78.1 \\ \small(±0.52) \end{tabular} & \begin{tabular}[c]{@{}l@{}} 66.7  \\ \small(±2.00) \end{tabular}\\
\arrayrulecolor[HTML]{D9D9D9}\hline
\textit{Full\_No-history\_Weighted} & \begin{tabular}[c]{@{}l@{}}66.1\\ \small(±1.33) \\ \tiny{*}6.27e-04\end{tabular} &  \begin{tabular}[c]{@{}l@{}} 65.7 \\ \small(±0.57) \end{tabular} &   \begin{tabular}[c]{@{}l@{}} 66.4 \\ \small(±0.55)\end{tabular}&  \begin{tabular}[c]{@{}l@{}} 58.3 \\ \small(±2.3) \end{tabular}\\
\arrayrulecolor[HTML]{D9D9D9}\hline
\textit{Full\_History\_Non-weighted} & \begin{tabular}[c]{@{}l@{}}73.4\\  \small(±1.01) \\ \tiny{*}3.78e-02 \end{tabular} & \begin{tabular}[c]{@{}l@{}} 72.9 \\ \small(±0.51) \end{tabular} & \begin{tabular}[c]{@{}l@{}} 75.0 \\ \small(±0.52) \end{tabular}& \begin{tabular}[c]{@{}l@{}} 50.0 \\ \small(±2.23) \end{tabular}\\
\arrayrulecolor[HTML]{D9D9D9}\hline
\textit{Full\_No-history\_Non-weighted} & \begin{tabular}[c]{@{}l@{}}71.1\\ \small(±0.99) \\ \tiny{*}2.75e-03\end{tabular} & \begin{tabular}[c]{@{}l@{}}72.1 \\ \small(±0.55)  \end{tabular}& \begin{tabular}[c]{@{}l@{}}71.9 \\ \small(±0.50) \end{tabular}& \begin{tabular}[c]{@{}l@{}}75.0 \\ \small(±1.9) \end{tabular}\\
\bottomrule
\end{tabular}}
\begin{tablenotes}
\small
\item[$^{*}$] indicates the p-value w.r.t. the History\_weighted model
\end{tablenotes}
\caption{Test set performance metrics of models trained on the Stanford TCE dataset. The metrics reported are ROC AUC score (AUC), Accuracy (Acc), Recall (Rec) and Specificity (Spec) in percentage (\%).}
\label{tab:res_tce_stanf}
\end{table}

The Stanford TCE dataset was employed in two distinct ways:
\begin{itemize}
    \item As an external validation set for testing models already trained on the EIDB-derived datasets. The results are shown in Table \ref{tab:res_ext_val}. Although this analysis does not yield results as conclusive as those obtained on the EIDB-derived test sets presented in Table 2 of the main paper, they point the same direction. The loss in the model performance is mainly due to the unbalanced distribution of the TCE dataset, which is skewed toward cases of treatment failure, unlike the dataset used for model training. Nevertheless, the analysis confirms that incorporating the temporal dynamics of mutations improves prediction accuracy compared with the standard analysis of the last available genotype, helping more in the prediction of failures than successes.  
    \item As a new dataset to be divided into training set and test set to train and test the \textit{Full} models. The results are shown in Table \ref{tab:res_tce_stanf}. The statistically significant differences between the AUC of the models compared to the \textit{Full\_History\_Weighted\_model} confirm the importance of considering the mutation history of each patient when evaluating the administration of a new ART. 
\end{itemize}

\end{document}